\documentclass[10pt,twocolumn,letterpaper]{article}

\usepackage[english]{babel}
\usepackage[pagenumbers]{cvpr} 

\usepackage[accsupp]{axessibility} 

\usepackage{amsmath}
\usepackage{amssymb}
\usepackage{mathtools}
\usepackage{makecell}
\usepackage{svg}
\usepackage[T1]{fontenc}
\usepackage{newtxtext,newtxmath}
\usepackage{bm}
\usepackage{iftex}

\usepackage{nicefrac}
\usepackage[scale=0.85]{tgcursor}
\usepackage[acronym]{glossaries}

\usepackage{enumitem}
\usepackage{graphicx}
\usepackage{booktabs}
\usepackage{microtype}
\usepackage[skins]{tcolorbox}

\usepackage{dblfloatfix}
\usepackage{placeins}
\usepackage{multirow}

\definecolor{cvprblue}{rgb}{0.21,0.49,0.74}
\usepackage[pagebackref,breaklinks,colorlinks,citecolor=cvprblue]{hyperref}
\usepackage[capitalize]{cleveref}
\crefname{section}{Sec.}{Secs.}
\Crefname{section}{Section}{Sections}
\Crefname{table}{Table}{Tables}
\crefname{table}{Tab.}{Tabs.}

\def\cvprPaperID{20} 
\def\confName{CVPR}
\def\confYear{2024}

\setlist[itemize,enumerate]{left=0.1ex,
  itemsep=0.5ex,
  partopsep=0pt,
  topsep=1ex,
  parsep=0pt,
  labelsep=0.5ex
}

\newacronym{sol}{SOL}{String Object List}
\newacronym{nlp}{NLP}{Natural Language Processing}
\newacronym{gnn}{GNN}{Graph Neural Network}
\newacronym{gan}{GAN}{Graph Analysis Network}
\newacronym{gatn}{GAtN}{Graph Attention Network}
\newacronym{sa}{SA}{Self Attention}
\newacronym{msa}{MSA}{Multihead Self Attention}
\newacronym{ga}{GA}{Graph Attention}
\newacronym{ssl}{SSL}{Self-Supervised Learning}
\newacronym{kl}{KL}{Kullback-Leibler}

\newacronym{mdct}{MDCT}{Modified Discrete Cosine Transform}
\newacronym{ssast}{SSAST}{Self-Supervised Audio Spectrogram  Transformer}
\newacronym{cbr}{CBR}{Constant Bit Rate}
\newacronym{cct}{CCT}{Compact Convolutional Transformer}
\newacronym{tssdnet}{TSSDNet}{Time-Domain Synthetic Speech  Detection Network}
\newacronym{lfccs}{LFCCs}{Linear Frequency Cepstral Coefficients}
\newacronym{mfccs}{MFCCs}{Mel Frequency Cepstral Coefficients}
\newacronym{cqccs}{CQCCs}{Constant Q Cepstral Coefficients}
\newacronym{cfccs}{CFCCs}{Cochlear Filter Cepstral Coefficients}
\newacronym{svm}{SVM}{Support Vector Machine}
\newacronym{gmms}{GMMs}{Gaussian Mixture Models}
\newacronym{ws}{WS}{Windowing Scheme}
\newacronym{sf}{SF}{scale factors}
\newacronym{cnn}{CNN}{Convolutional Neural Network}
\newacronym{hvxc}{HVXC}{Harmonic Vector Excitation Coding}
\newacronym{auprc}{AUPRC}{Area Under Precision Recall Curve}
\newacronym{lcnn}{LCNN}{Light Convolutional
Neural Network}
\newacronym{blstm}{BLSTM}{Bidirectional Long Short-Term Memory}
\newacronym{lstm}{LSTM}{Long Short-Term Memory}
\newacronym{rcnn}{RCNN}{Residual Convolutional
Neural Network}
\newacronym{rnn}{RNN}{Recurrent Neural Network}
\newacronym{m2d4speech}{M2D4Speech}{Masked Modeling Duo for Speech}
\newacronym{m2d}{M2D}{Masked Modeling Duo}
\newacronym{eer}{EER}{Equal Error Rate}
\newacronym{tts}{TTS}{Text-to-Speech}
\newacronym{fnr}{FNR}{False Negative Rate}
\newacronym{fpr}{FPR}{False Positive Rate}
\newacronym{vad}{VAD}{Voice Activity Detection}

\newacronym{qmf}{QMF}{Quadrature Mirror Filter}
\newacronym{imdct}{IMDCT}{Inverse Modified Discrete Cosine Transform}
\newacronym{mfcc}{MFCC}{Mel-Frequency Cepstrum Coefficients}
\newacronym{cqt}{CQT}{Constant-Q Transform}
\newacronym{stft}{STFT}{Short Time Fourier Transform}
\newacronym{dft}{DFT}{Discrete Fourier Transform}
\newacronym{dct}{DCT}{Discrete Cosine Transform}
\newacronym{fft}{FFT}{Fast Fourier Transform}
\newacronym{auc}{AUC}{Area Under Curve}
\newacronym{roc}{ROC}{Receiver Operating Characteristics}
\newacronym{aac}{AAC}{Advanced Audio Coding}
\newacronym{flac}{FLAC}{Free Lossless Audio Codec}
\newacronym{mlp}{MLP}{Multi Layer Perceptron Network}
\newacronym{bce}{BCE}{Binary Cross Entropy}
\newacronym{rocauc}{ROC-AUC}{Area Under the Receiver Operating Characteristic Curve}
\newacronym{passt}{PaSST}{Patchout faSt Spectrogram Transformer}

\newacronym{cqost}{CQ-OST}{Constant-Q Octave Subband Transform}
\newacronym{icqccs}{ICQCCs}{Inverted Constant-Q Cepstral Coefficients}
\newacronym{dnn}{DNN}{Deep Neural Network}
\newacronym{se}{SE}{Squeeze-and-Excitation}
\newacronym{bilstm}{Bi-LSTM}{Bidirectional Long Short-Term Memory}
\newacronym{resnet}{ResNet}{Residual Neural Network}
\newacronym{vaes}{VAEs}{Variational Autoencoders}
\newacronym{asr}{ASR}{Automated Speaker Recognition}

\glsdisablehyper

\ifxetex\relax
\else
\DeclareUnicodeCharacter{2212}{$-$}
\fi

\ExplSyntaxOn

\prop_new:N \g_pretrain_lut_prop
\prop_gset_from_keyval:Nn \g_pretrain_lut_prop {
  DC=DC,
  TC=TC,
  IC=IC,
  NR=NR,
  LR=LR
}

\newcommand{\pretaskone}[1]{
  \tl_set:Nx \l_tmpb_tl {\str_uppercase:n {#1}}
  \prop_get:NVNF \g_pretrain_lut_prop \l_tmpb_tl \l_tmpa_tl {
    \GenericError{}{invalid~pretrain~task~#1}{}{}
  }

  \raisebox{-0.3ex}{
    \tcbox[
    enhanced,
    standard~jigsaw,
    opacityback=0,
    colback=white,
    colframe=black,
    boxrule=0.5pt,
    left=0mm,
    right=0mm,
    top=0mm,
    bottom=0mm,
    boxsep=1.5pt
    ]{\parbox{1.5em}{\centering \l_tmpa_tl}}
  }
}

\newcommand{\pretask}[1]{
  \clist_map_inline:nn {#1} {
    \pretaskone{##1}
  }
}

\ExplSyntaxOff

\begin{document}

\setlength{\abovedisplayskip}{4pt}
\setlength{\belowdisplayskip}{4pt}
\setlength{\abovedisplayshortskip}{2pt}
\setlength{\belowdisplayshortskip}{2pt}

\newcommand{\td}{$^\dagger$}
\newcommand{\tdd}{$^\ddagger$}
\title{FairSSD: Understanding Bias in Synthetic Speech Detectors}
\author{
\parbox{0.95\linewidth}{
\hspace*{\fill} Amit Kumar Singh Yadav\td \hfill Kratika Bhagtani\td \hfill Davide Salvi\tdd \hfill Paolo Bestagini\tdd \hfill Edward J. Delp\td \hspace*{\fill}\\
\vspace*{0.1em}\\
    \small\centering \td  Video and Image Processing Lab (VIPER), Purdue University, West Lafayette, Indiana, USA\\
    \small\centering \tdd Dipartimento di Elettronica, Informazione e Bioingegneria, Politecnico di Milano, Milano, Italy
}
}

\maketitle
\begin{abstract}
Methods that can generate synthetic speech which is perceptually indistinguishable from speech recorded by a human speaker, are easily available.
Several incidents report misuse of synthetic speech generated from these methods to commit fraud.
To counter such misuse, many methods have been proposed to detect synthetic speech. 
Some of these detectors are more interpretable, can generalize to detect synthetic speech in the wild
and are robust to noise. 
However, limited work has been done on understanding bias in 
these detectors.
In this work, we examine bias in existing synthetic speech detectors to determine if
they will unfairly target a particular gender, age and accent group.
We also inspect whether these detectors will have a higher misclassification rate for bona fide speech from speech-impaired speakers w.r.t fluent speakers. 
Extensive experiments on 6 existing synthetic speech detectors using more than 0.9 million speech signals demonstrate that most detectors are gender, age and accent biased, and future work is needed to ensure fairness.
To support future research, we release our evaluation dataset, models used in our study and source code at \url{https://gitlab.com/viper-purdue/fairssd}.
%
\end{abstract}


\section{Introduction}
Speech signals can be bona fide (\ie, recorded from a human speaker) or synthetic (\ie, 
generated from a computer)~\cite{asvspoof19, bhagtani2022overview}.
With development in 
Generative Artificial Intelligence (AI)~\cite{ho2020, dhariwal2021diffusion, stable_diffusion_cvpr2022}, methods to generate high-quality synthetic speech are easy available~\cite{ elevenlabs2023,unitspeech_interspeech2023,xttsv2, huang2022prodiff}.
Many of these methods can mimic characteristics of an individual's voice, for example, its patterns, intonations, and pronunciations, which makes the synthesized speech
perceptually indistinguishable from the recorded real one.
Several recent incidents have reported misuse of such high-quality synthetic speech for spreading misinformation and being able to commit financial fraud~\cite{nyt_2023, bi_2023, fortune_2023, wp_2023}.

\begin{figure}[!t]
    \centering
    \includegraphics[width=1.0\linewidth]{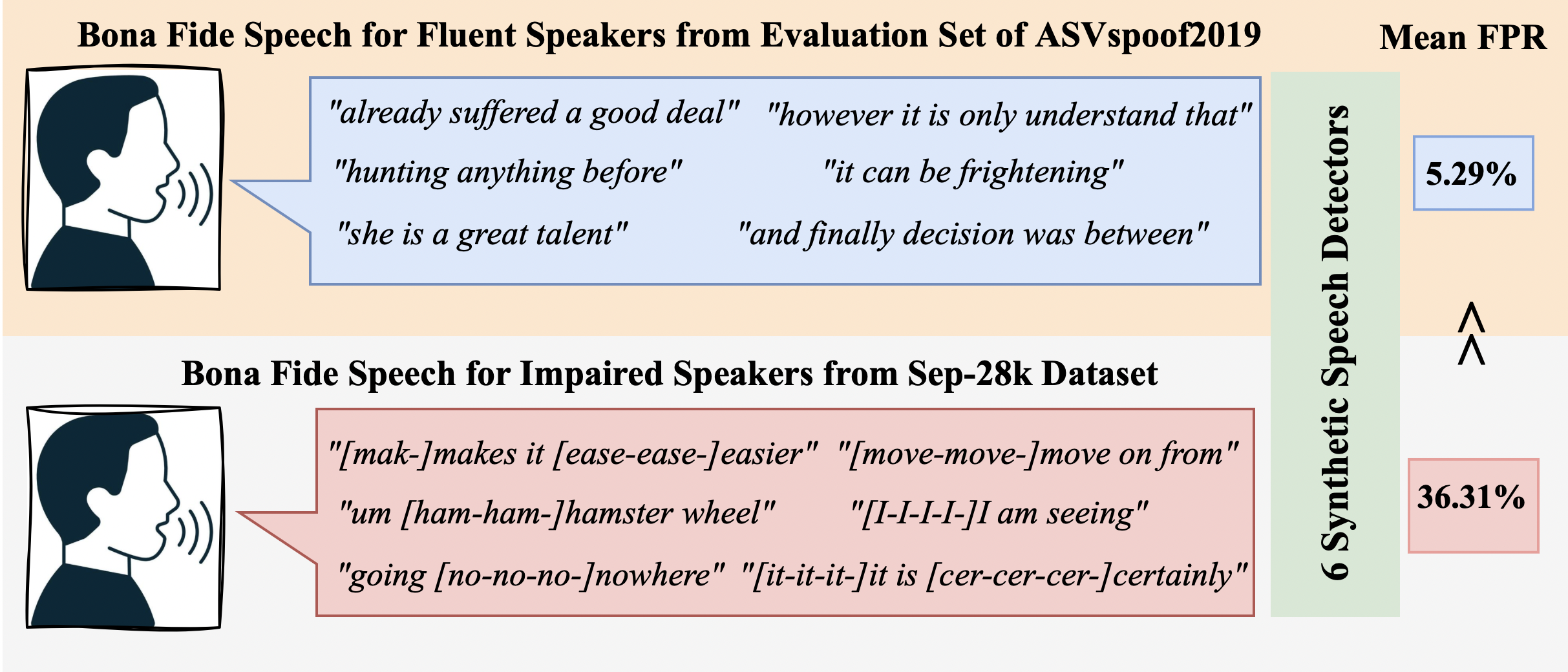}
    \caption{Mean False Positive Rate (FPR) of 6 synthetic speech detectors on bona fide speech from fluent and speech-impaired speakers.}
    \label{fig:fpr}
\end{figure}

To prevent the malicious use of synthetic speech, existing works have proposed several synthetic speech detectors~\cite{tssdnet_2021, asvspoof_2021, tak22_odyssey, add2023_submission_li2022convolutional, acm_21_logspec, Sun_2023_CVPR, alan_acm_2023}.
Moreover, the multimedia forensics community has organized challenges and released datasets, such as ASVspoof2019~\cite{asvdata_2019}, 
to foster research and progress in this domain.
Some of the released detectors focus on detecting synthetic speech in in-the-wild conditions (evaluation on speech synthesized from generation methods not seen before during training)~\cite{in_the_wild, Sun_2023_CVPR, ps3dt}.
Other methods aim to produce interpretable results~\cite{dsvae_arXiv, exp_fake_22, exp_cqcs_slt20, salvi2023towards, yadav2023dsvae}.
Finally, other methods are claimed as robust to compression~\cite{asvspoof_2021, ps3dt, assd_2023}.
Despite the great efforts of multimedia forensic researchers, there is very limited work in understanding bias in these synthetic speech detectors.
By bias, we refer to an action in which a detector unfairly targets a specific demographic group of individuals and falsely labels their bona fide speech as synthetic. 
~\cref{fig:fpr} shows one such bias condition from our study in which the mean misclassification rate (\gls{fpr}) for all detectors is higher on bona fide speech from speech-impaired speakers than on bona fide speech from fluent speakers.

Several existing works have examined bias in~\gls{asr} systems and deepfake face detectors~\cite{bias_asv_interspeech_2021, acm_bias_asv_2022, asr_inclusive_2022, xu2023comprehensive}.
For example, Pu~\etal~\cite{deepfake_fair_23} observed that a deepfake face detector performs better for females as compared to males. 
Similar bias with respect to gender, age, speech impairment, race, and accent has also been shown in~\gls{asr} systems~\cite{bias_asv_interspeech_2021}.
However, limited attempts have been made in understanding bias in synthetic speech detectors.
Ensuring fairness in synthetic speech detectors 
is important to 
prevent any misclassifications of bona fide speech from a particular ethnic and demographic group as synthetic, when 
these detectors are deployed on social platforms.
As this can impact public opinion, it can lead to unintended and significant societal and political consequences, reducing trust on synthetic speech detectors and eroding reputation of the social platform using biased detectors.

In this work, we address this issue by thoroughly investigating and analyzing bias in synthetic speech detectors. 
We examine bias in six different approaches for detecting synthetic speech shown in~\cref{fig:overview} and described in~\cref{sec:detectors}.
In our first 3 experiments, we examine whether synthetic speech detectors target a particular gender, age and accent group.
We processed Mozilla Common Voice Corpora~\cite{mozilla_cvc} and obtained approximately 0.9 million bona fide speech signals for our first 3 experiments.
In our last experiment, we examine if synthetic speech detectors misclassify bona fide speech from people with speech impairments such as stuttering. 
We used bona fide speech from Sep-28K~\cite{sep_1, sep_2} dataset for this study.
Our results reveal hidden bias in existing synthetic speech detectors and we believe it will bring attention of forensics community to address these biases.
\section{Related Work}\label{sec:related-work}
In this section, we describe methods for synthetic speech detection and existing work on the fairness of forensics detectors.

\subsection{Synthetic Speech Detection}\label{sec:detectors}
\begin{figure}[!t]
    \centering
    \includesvg[width=1.0\linewidth]{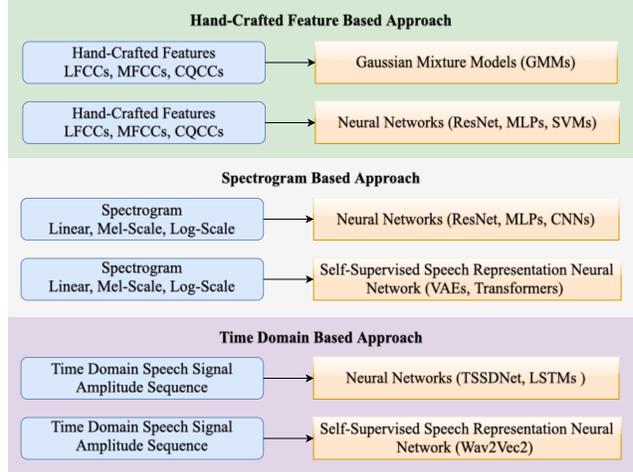}
    \caption{Existing Approaches for Synthetic Speech Detection.}
    \label{fig:overview}
\end{figure}

An overview diagram of existing approaches for detecting synthetic speech is shown in~\cref{fig:overview}.
Based on the input, synthetic speech detection methods can be split into three categories.

The first and most conventional category of approaches obtains hand-crafted features such as ~\gls{mfccs}~\cite{mfccs},~\gls{cqccs}~\cite{cqccs}, bit-rate~\cite{Borzi_2022_CVPR}, and ~\gls{lfccs}~\cite{li2021replay} from the speech signal to detect synthetic speech. 
Some methods process the obtained hand-crafted features using~\gls{gmms}~\cite{lfcc_interspeech, asvspoof19} and others use deep neural networks such as \gls{bilstm}~\cite{akdeniz2021detection} and \gls{resnet}~\cite{alzantot2019, he2016deep}.

The second category of approaches involves processing spectrogram images of a speech signal~\cite{add2023_submission_li2022convolutional, spec_vgg_sincnet_28, ps3dt, bartusiak2023, acm_21_logspec}.
The spectrogram is a 2D representation that plots temporal variations in magnitude of different frequency components of the speech signal~\cite{mel}.
The frequency scale can be linear~\cite{spec_vgg_sincnet_28}, logarithmic~\cite{acm_21_logspec, alzantot2019} or it can be based on mel-filters~\cite{mel, bartusiak2023}.
These methods then process the spectrogram using \gls{cnn} and \gls{rnn}~\cite{add2023_submission_li2022convolutional, spec_vgg_sincnet_28}.
Some recent methods in this approach use self-supervised learning and train networks such as \gls{vaes}~\cite{vae_first_paper_2013} or a transformer neural network~\cite{vaswani_2017} on millions of real speech signals to first obtain a general speech representation network~\cite{koutini_2021, gong_2021_ssast, gong_2021_ast} and then fine-tune that network on limited training data for synthetic speech detection~\cite{ps3dt, bartusiak2023, yadav2023dsvae}.
Such approaches are also effective in localizing the synthetic region and attributing the synthesizer used for generating it~\cite{amit_ei_paper, acm_kratika_2023, mdrt_amit, fgsat_2023}

Methods in the third category process the time domain representation of the speech signal~\cite{tssdnet_2021, multi_task_zhang21_asvspoof, Sun_2023_CVPR, acm_local_attention_paper_2023}. 
By time domain speech signal, we refer to 1D sequence that corresponds to temporal change in the amplitude of the speech signal.
The classification network can either be a neural network trained from scratch \eg, \gls{tssdnet} and \gls{lstm} network as in~\cite{tssdnet_2021,
multi_task_zhang21_asvspoof, acm_local_attention_paper_2023, Sun_2023_CVPR} or a pre-trained general speech representation neural network such as wav2vec 2.0~\cite{wav2vec2_2020, PartialSpoof, tak22_odyssey}. 
Most of the methods use AudioSet~\cite{audioset} or Libri-Speech dataset~\cite{librispeech} for any pre-training to obtain general speech representation neural network~\cite{gong_2021_ssast, gong_2021_ast, koutini_2021, wav2vec2_2020}.
For fine-tuning or from-scratch training and evaluation almost all the methods use ASVspoof2019~\cite{asvdata_2019} dataset.

\subsection{Fairness of Forensic Detectors}
Several existing works examine and improve generalization capability~\cite{asvspoof_2021, tak22_odyssey, in_the_wild, Sun_2023_CVPR}, robustness to compression~\cite{assd_2023, ps3dt}, and explainability~\cite{exp_fake_det_icassp22,yadav2023dsvae} of synthetic speech detectors. 
Some work has been done in understanding bias in~\gls{asr} systems~\cite{bias_asv_interspeech_2021, acm_bias_asv_2022, asr_inclusive_2022}.
Many works have explored bias in face recognition systems~\cite{wacv_21_fairface, 21_survey_bias_fairness, 22_tbiom_ai_fair}.
These have been extended to evaluate bias in deepfake face detectors~\cite{deepfake_fair_23, open_challenge_23, wacv_w_22, fairness_face_2023, xu2023comprehensive, trinh2021examination}.
Trinh~\etal~\cite{trinh2021examination} examined bias in three existing deepfake face detectors and observed a disparity in performance of more than 10 percentage points between different demographic groups. Hazirbas~\etal~\cite{22_tbiom_ai_fair} had similar observations.
Pu~\etal~\cite{deepfake_fair_23} evaluated a popular deepfake face detector 
and observed that it performs better for female gender.
Xu~\etal~\cite{xu2023comprehensive} annotated a deepfake detection dataset and performed a thorough bias evaluation of deepfake face detectors. 
Nadimpalli~\etal~\cite{nadimpalli2022gbdf}
observed gender bias in deepfake face detectors and worked on reducing it by training on a more gender-balanced dataset. However, such an approach requires time-consuming data annotation. 
To counter this, Ju~\etal~\cite{Ju_2024_WACV} presented a loss function for improving fairness of deepfake detectors.
This loss function can work in both scenarios, \ie, when demographic annotations are either available or absent during training.
While many initial works have analyzed and showed evidence of biases in deepfake face detectors, reducing those biases is still an open research area~\cite{open_challenge_23}.
In this work, similar to previous work examining biases in deepfake face detectors, we attempt to understand biases in synthetic speech detectors.
\section{Proposed Study}\label{sec:method}
In this section, we describe the synthetic speech detectors examined in this work, the dataset used for training them, the evaluation datasets and metrics used in our bias study.

\subsection{Detectors Used in Our Study}\label{sec:detector_used}
An overview of different categories of synthetic speech detection methods
is provided in~\cref{fig:overview} and~\cref{sec:detectors}.
To run a comprehensive study, we include methods from each category shown in \cref{fig:overview}.
In total, we examine 6 different synthetic speech detection methods.
Out of these 6 methods, we trained 4 of them from scratch to obtain detectors for our study. 
For the remaining 2, we perform our study directly on the model and weights released by the authors. 
More details about each method and its training setup are presented below.\\
\textbf{LFCC-GMMs~\cite{asvspoof19, lfcc_interspeech}} : 
We use this method as a representative of methods that process hand-crafted features using \gls{gmms}.
It is the best hand-crafted baseline provided in the ASVspoof2019~\cite{asvdata_2019} challenge.
In this method, \gls{lfccs}~\cite{lfcc_interspeech} are obtained from the speech signal.
After obtaining the \gls{lfccs} features, for each class (bona fide speech class and synthetic speech class), we fit a separate GMM for each class.
Each GMM estimates the probability distribution of the features for that class.
During evaluation, we obtain \gls{lfccs} from a given speech signal and use \gls{gmms} to obtain the probability of it belonging to each class.
The speech signal is labeled as the most probable class.
We performed an ablation study to determine the best set of hyper-parameters for obtaining the hand-crafted features. More details about the ablation study and training can be found in the supplement material. 
The model used in this study processes 30ms windows with a hop size of 15ms and frequency components up to 4KHz to obtain \gls{lfccs} features.
GMM for each class has 512 Gaussian mixture components.\\
\textbf{MFCC-ResNet~\cite{alzantot2019} :}
We use this method as a representative of the methods that obtain hand-crafted features and process them using a neural network.
For this method, we trained and obtained our own model weights.
This method uses~\gls{mfcc}~\cite{mfccs}, which were also used in baselines provided in the ASVspoof2019 challenge~\cite{asvspoof19}.
The~\gls{mfcc} are obtained using the~\gls{stft} of the speech signal, mel-spectrum filters and ~\gls{dct}~\cite{alzantot2019}.
For our experiments, we select the first 24~\gls{mfcc} coefficients.
Besides~\gls{mfcc}, we also obtain the first and second derivatives of~\gls{mfcc}.
Overall, we use features of dimension 72.
These hand-crafted features are processed by a ResNet~\cite{he2016deep} for synthetic speech detection.\\
\textbf{Spec-ResNet~\cite{alzantot2019} :}
We use this method as a representative for methods that process spectrograms using a neural network.
For this method, we trained and obtained our own model weights.
This method obtains the spectrogram by taking~\gls{stft} of the input speech and squaring its magnitude. 
The scale is changed to a logarithmic scale.
A ResNet network~\cite{he2016deep} processes the logarithmic-scaled spectrogram for synthetic speech detection.\\
\textbf{PS3DT~\cite{ps3dt} :} 
This method is representative of methods that process spectrogram representation of the speech signal using a self-supervised pre-trained network.
We implemented and trained this method.
We selected this method as it performs better than others which use a similar approach~\cite{bartusiak2023} and process spectrograms.
It also has better in-the-wild performance and is more robust to compression~\cite{ps3dt}.
It uses mel-scale~\cite{mel} spectrogram.
The network is pre-trained on the Audioset Dataset~\cite{audioset} using self-supervised learning.
PS3DT divides a mel-spectrogram into patches and obtains patch representations using a transformer neural network.
During pre-training, some patches are masked and reconstructed~\cite{msm_mae_niizumi}.
Later, the pre-trained network is fine-tuned on ASVspoof2019 training set~\cite{asvdata_2019}.
During fine-tuning, the patch representations are rearranged such that representations corresponding to the same temporal location are aligned together.
A~\gls{mlp} process these representations to label speech as bona fide or synthetic.\\
\textbf{TSSDNet~\cite{tssdnet_2021} :} 
This method is representative of an end-to-end method that processes time-domain speech signals.
We used the weights released by authors in~\cite{tssdnet_2021}.
We used this method as it is shown in ~\cite{aasist, tssdnet_2021} that it performs better than other similar time-domain methods such as Raw PC-DARTS~\cite{rawpc_darts} and is also found robust to compression noise~\cite{assd_2023}.
In \gls{tssdnet}, two kinds of networks are proposed, namely, ResNet style and Inception style networks, which are used to process speech amplitude~\cite{tssdnet_2021}. 
In~\cite{tssdnet_2021}, the ResNet version performs better than the Inception version, and therefore we use the ResNet version in our experiments.\\
\textbf{Wav2Vec2-AASIST~\cite{tak22_odyssey}:}
This method is representative of methods that process time-domain speech signals using a large self-supervised neural network. 
We used the weights provided by the authors in~\cite{tak22_odyssey}.
We use this method as it has nearly perfect performance on ASVspoof2019~\cite{asvdata_2019} and has better generalization performance than several existing methods on the ASVspoof2021 dataset~\cite{tak22_odyssey}. 
This method processes the time domain speech signal using a pre-trained wav2vec2 network~\cite{wav2vec2_2020} and combines it with spectral features using the Audio Anti-Spoofing using Integrated Spectro-Temporal (AASIST) graph attention network~\cite{aasist}. 
Wav2vec2 has more than 300 million parameters and is trained on a large audio dataset using self-supervision. 
It consists of a~\gls{cnn} and a transfomer network~\cite{tak22_odyssey, vaswani_2017}.

\subsection{Datasets Used in Our Study}\label{sec:dataset}
\begin{table}[!t]
    \centering
    \caption{Details of speech signals in bona fide class of each evaluation set used for gender, accent and age bias studies.}
    \resizebox{0.9\columnwidth}{!}{%
\begin{tabular}{@{\extracolsep{-4pt}}lccccc}
\toprule
\makecell{\bfseries Study} &
\makecell{\bfseries Set Name} & 
\makecell{\bfseries Accent} & \makecell{\bfseries Age Group} & \makecell{\bfseries Gender} & \makecell{\bfseries Samples} \\ 
\midrule
\multirow{6}{*}{Gender} & $D_{US-20s-M}$ & US English & 20s & Male & 31,000 \\
& $D_{US-20s-F}$ & US English & 20s & Female & 31,000 \\
& $D_{US-30s-M}$ & US English & 30s & Male & 15,000 \\
& $D_{US-30s-F}$ & US English & 30s & Female & 15,000 \\
& $D_{US-60s-M}$ & US English & 60s & Male & 16,000 \\
& $D_{US-60s-F}$ & US English & 60s & Female & 16,000 \\ 
\midrule
\multirow{12}{*}{Age} & $D_{US-ts-F}$ & US English & teens & Female & 8,900 \\
 & $D_{US-20s-F}$ & US English & 20s & Female & 8,900 \\
 & $D_{US-30s-F}$ & US English & 30s & Female & 8,900 \\
 & $D_{US-40s-F}$ & US English & 40s & Female & 8,900 \\
 & $D_{US-50s-F}$ & US English & 50s & Female & 8,900 \\
 & $D_{US-60s-F}$ & US English & 60s & Female & 8,900 \\
 & $D_{US-ts-M}$ & US English & teens & Male & 8,900 \\
 & $D_{US-20s-M}$ & US English & 20s & Male & 8,900 \\
 & $D_{US-30s-M}$ & US English & 30s & Male & 8,900 \\
 & $D_{US-40s-M}$ & US English & 40s & Male & 8,900 \\
 & $D_{US-50s-M}$ & US English & 50s & Male & 8,900 \\
 & $D_{US-60s-M}$ & US English & 60s & Male & 8,900 \\
 
\midrule

\multirow{10}{*}{Accent} & $D_{US-20s-F}$ & US English & 20s & Female & 4,900 \\
& $D_{SA-20s-F}$ & South Asian & 20s & Female & 4,900 \\
& $D_{CN-20s-F}$ & Canadian & 20s & Female & 4,900\\
& $D_{UK-20s-F}$ & British & 20s & Female & 4,900\\
& $D_{AU-20s-F}$ & Australian & 20s & Female & 4,900\\
& $D_{US-20s-M}$ & US English & 20s & Male & 8,100\\
& $D_{SA-20s-M}$ & South Asian & 20s & Male & 8,100\\
& $D_{CN-20s-M}$ & Canadian & 20s & Male & 8,100 \\
& $D_{UK-20s-M}$ & British & 20s & Male & 8,100\\
& $D_{AU-20s-M}$ & Australian & 20s & Male & 8,100\\

 \bottomrule
\end{tabular}}

    \label{tab:dataset}
\end{table}
In this section, we first describe the dataset used for training the detectors and evaluating their detection performance. 
We later discuss our evaluation dataset prepared for examining the age, gender and accent bias in each detector.
We also discuss the dataset used for the bias study on speech with stuttering impairment.
\subsubsection{Detection Training and Evaluation Dataset}
Following most existing work on synthetic speech detection~\cite{tssdnet_2021, tak22_odyssey, asvspoof_2021}, we use Logical Access (LA) part of the ASVSpoof2019 Dataset~\cite{asvdata_2019} for training.
Each of the existing methods is trained, validated and evaluated on the official training set ($D_{tr}$), development set ($D_{dev}$) and evaluation set ($D_{eval}$), respectively.
There are 25,380 speech signals in $D_{tr}$, 24,844 in $D_{dev}$, and 71,237 speech signals in $D_{eval}$ out of which 2580, 2548, and 7355, respectively, are bona fide speech signals.
The synthetic speech signal in $D_{tr}$ and $D_{dev}$ are generated using 6 different synthetic speech generators. 
The synthetic speech signals in $D_{eval}$ are generated from 13 different speech generators, 2 generators are the same as the ones used in $D_{tr}$ and $D_{dev}$ sets, while 11 generators are unknown generators not present in $D_{tr}$ and $D_{dev}$ sets.
Hence, the evaluation set has a majority of synthetic speech signals generated from unknown speech synthesizers not seen during training.
Speakers recorded for bona fide speech do not overlap among any sets and the sampling rate is 16 kHz.
In all our bias studies, we kept the same sampling rate during evaluation.
\subsubsection{Evaluation Datasets for Bias Study}
In our study, we want to examine if existing synthetic speech detectors will unfairly target bona fide speech from a particular group and misclassify it as synthetic. 
To facilitate our first 3 studies which examine gender, age and accent bias, we need bona fide speech signals with demographic annotations such as gender, age, and accent.
We use bona fide speech samples from multi-language speech corpus Mozilla Common Voice Corpus 16.1~\cite{mozilla_cvc}.
We use only the validated and the English subset of the Mozilla CVC dataset~\cite{mozilla_cvc} that has approximately 1.78 million speech signals, in total constituting 2,600 hours of bona fide speech.
Many of these speech signals do not have all the required annotations for our study.
We pre-processed the dataset and obtained approximately 0.9 million speech signals having 
required annotations.
From these, we constructed 28 evaluation sets for performing our bias study.
All 28 evaluation sets consist of speech signals from bonafide class as well as synthetic class. 

First, we describe the details of speech from bonafide class, which comes from Mozilla CVC dataset~\cite{mozilla_cvc}.
The number of bona fide signals in each evaluation set is reported in ~\cref{tab:dataset}. 
We selected the numbers based on the minimum number of samples present in one group, which is explained next.
For example, in our gender bias study with speakers in age group $20s$ that have US accent, we have two sets, namely $D_{US-20s-M}$ and $D_{US-20s-F}$ as shown in~\cref{tab:dataset}.
The biggest set with demographic: female, US accent and in age group 20s has 31.5K bona fide speech signals and similarly the biggest set with demographic: male, US accent and in age group 20s has 110K bona fide speech signals.
We randomly sample 31K speech signals from each of the male and female sets because this is the approximate minimum of both numbers.
For consistency, we need to select equal number of speech signals from male and female genders.
We repeated this random sampling 5 times, leading to 5 versions of each set mentioned in~\cref{tab:dataset}. 
Therefore, each experiment is run 5 times and the mean and standard deviation for each performance metric is provided in all of our bias studies.

Next, we describe the details of the synthetic class of all 28 evaluation sets.
The synthetic speech class is same in each set and it consists of all the synthetic speech signals from the $D_{eval}$ set of ASVspoof2019 Dataset.
This is done so that any difference in the performance of a detector on two different sets can directly be attributed to the difference in its performance on demographic of bonafide class in the sets.
We created 6 sets for our gender bias study. 
We fixed accent to US English and evaluated for 3 different ages (see~\cref{tab:dataset}).
For each age, we created two sets, one with bona fide speech from only male speakers and other with only female speakers.
Similarly, we created 10 sets for understanding accent bias and 12 sets for examining age bias.
More details about the dataset and pre-processing are provided in the supplement.

For examining bias on stuttering speech, we use bona fide speech from SEP-28k dataset~\cite{sep_1, sep_2} and synthetic speech from $D_{eval}$ set of ASVspoof2019 dataset.
It contains approximately 28K bona fide speech signals.
The stuttering speech can have prolongation: elongated syllable (\eg, M[mmm]ommy), gasps for air or stuttered pauses, repeated syllables (\eg, I [pr-pr-pr-]prepared dinner), word or phrase repetition (\eg, I made [made] dinner), and filler words use to cope with stutter (\eg, "um" or "uh").
We pre-processed the bona fide speech to remove speech samples with poor audio quality, music in background, and signals with just silence or background noise.
After pre-processing, we obtained 21,855 bona fide stuttering speech signals.

\subsection{Evaluation Metrics}\label{sec:metrics}
We use two metrics for evaluation. 
First, we use \gls{eer} for measuring performance of each detector on ASVspoof2019 dataset.
\gls{eer} is the performance metric used in ASVspoof2019 challenge~\cite{asvspoof19}.
For calculating \gls{eer}, a decision threshold is determined that balances and makes both \gls{fpr} and \gls{fnr} equal. 
The \gls{eer} is same as the \gls{fpr} or \gls{fnr} obtained at that threshold.
We obtain~\gls{eer} on ASVspoof2019 evaluation set ($D_{eval}$) as a performance measure of an individual synthetic speech detector.
Lower the \gls{eer}, the better the performance of the detector.
\gls{eer} is also used in existing speech recognition work for bias study~\cite{bias_asv_interspeech_2021, acm_bias_asv_2022}.
Difference in \gls{eer} values by a detector on \eg male and female sets is a measure of the gender bias.
For example, in our gender bias study we use  $D_{US-20s-M}$ and $D_{US-20s-F}$ sets shown in ~\cref{tab:dataset}. 
The synthetic speech class of each set is same and both sets have same number of bona fide speech.
As shown in~\cref{tab:dataset}, the age and accent for samples in bona fide class 
is also the same. 
Hence, the two sets only differ in terms of bona fide speech gender.
Therefore, the difference in \gls{eer} performance by a detector on $D_{US-20s-M}$ and $D_{US-20s-F}$ will imply that the detector targets bona fide speech from one group unfairly and is biased w.r.t gender.
In all our bias studies such as gender bias study (\cref{tab:gender_bias}), for each evaluation set, we report $\Delta EER := EER-minEER$ as a bias measure, where $minEER$ is minimum $EER$ of a detector obtained for a particular demographic group.
For example, in our gender bias study for speakers in age group 20s, we use $D_{US-20s-M}$ and $D_{US-20s-F}$ sets from ~\cref{tab:dataset}.
We obtain \gls{eer} of a particular detector on both sets.
The minimum of the two obtained \gls{eer}s is $minEER$ for that detector and used to report $\Delta EER$ on $D_{US-20s-M}$ and $D_{US-20s-F}$ sets.
Note: $\Delta EER$ will always be positive.
It will be zero for either one of $D_{US-20s-M}$ or $D_{US-20s-F}$ as one of the $EER$ values will be same as $minEER$. 
The higher $\Delta EER$ for the other set, the more biased the detector is for that set (gender) and bona fide speech from that group has been more unfairly 
mislabeled as synthetic.
The absolute values of \gls{eer} and more details about calculating $\Delta EER$ are provided in the supplement material.

The \gls{eer} provides an idea of a detector performance under the assumption that the distribution of the data under analysis is coherent with the distribution of the evaluation set used for the tests.
This may not be true, particularly if the detector is deployed on some platform for real-world use-case.
Hence, we also report the \gls{fpr} of each detector on sets shown in ~\cref{tab:dataset}.
\gls{fpr} measures the percentage of bona fide speech signals falsely labeled as synthetic.
All existing methods provide the probability of a speech signal being synthetic.
Therefore, the decision to label a speech signal as synthetic requires comparing the obtained probability from each method with a threshold.
We obtain the threshold for calculating \gls{fpr} on each set shown in ~\cref{tab:dataset} using an independent dataset. 
We use the evaluation set ($D_{eval}$) of ASVspoof2019 to obtain the threshold.
Speech signals in $D_{eval}$ set are generated from unknown speech generators and bona fide speakers do not overlap with the training set, making it a reasonable dataset for obtaining the threshold.

We obtain three \gls{fpr}s representing three different 
constraints.
The threshold for $FPR_{1}$ is where the detector has equal \gls{fpr} and \gls{fnr} on $D_{eval}$ set of ASVspoof2019 dataset.
The threshold for $FPR_{2}$ is where the detector has 0.08\gls{fpr} on $D_{eval}$ set of ASVspoof2019 dataset.
The threshold for $FPR_{3}$ is where the detector has 0.08\gls{fnr} on $D_{eval}$ set of ASVspoof2019 dataset.
Similar to the difference in \gls{eer} values, the difference in\gls{fpr} values for an individual detector (\eg, male and female subsets) is a measure of bias (\eg, gender bias).
Hence, in all bias studies, similar to \gls{eer}, we report $\Delta FPR_1$, $\Delta FPR_2$, and $\Delta FPR_3$.
Each of them will be a positive value.
Higher the $\Delta FPR$, the more biased the detector is for that particular evaluation set (demographic).
The absolute values and more details about $\Delta FPR$s are provided in the supplement material. 

\section{Experiments and Results}\label{sec:result}
\begin{table}[!t]
    \centering
    \caption{EER (in \%) performance of detectors on ASVspoof2019.}
    \resizebox{\columnwidth}{!}{%
\begin{tabular}{@{\extracolsep{-4pt}}lccccc}
\toprule
\makecell{\bfseries Detector\\ \bfseries Number \\ \bfseries (DN)} & \makecell{\bfseries Name }&
\makecell{\bfseries Type} &
\makecell{\bfseries Parameters} & 
\makecell{\bfseries \Vec{$D_{dev}$}}& 
\makecell{\bfseries \Vec{$D_{eval}$}} \\ 
\midrule
D01~\cite{tssdnet_2021}& {TSSDNet} & Time-domain & 0.35M & 0.74 & 1.62 \\
D02~\cite{tak22_odyssey}& {Wav2Vec2-AASIST}  & Time-domain & 317M & 0.02 & 0.23 \\
\midrule
D03~\cite{alzantot2019}& {Spec-ResNet} & Log-spectrogram & 0.32M & 0.71 & 10.10 \\
D04~\cite{ps3dt} & {PS3DT} & Mel-spectrogram & 95M & 2.82 & 4.54 \\
\midrule
D05~\cite{asvspoof19}& {LFCC-GMMs} & Hand-crafted & 0.1M & 0.04 & 3.67  \\
D06~\cite{alzantot2019} & {MFCC-ResNet} & Hand-crafted & 0.26M & 6.52 & 11.58 \\
\cmidrule(lr){5-6}
& & & $\boldsymbol{Mean}$ & 1.81 & 5.29 \\
\bottomrule
\end{tabular}}
    \label{tab:results-asv2019}
\end{table}
In this section, we discuss the performance of all detectors on ASVspoof2019 Dataset and experimental results from our bias studies.
We used $D_{dev}$ and $D_{eval}$ sets of ASVspoof2019 Dataset for measuring detection performance.
For bias studies, we used the evaluation dataset 
described in ~\cref{sec:dataset}.
\subsection{Experiment 1: Detection Performance}
We evaluate all detectors on the development set ($D_{dev}$) and the evaluation set ($D_{eval}$) of the ASVspoof2019 Dataset. 
The goal of this experiment is to check that the selected detectors have been correctly trained and work well for synthetic speech detection.

The results of this evaluation are shown in~\cref{tab:results-asv2019}.
The performance on $D_{dev}$ is representative of closed-set performance as synthetic speech is generated from the same synthesizers as used in the training set.
The performance on $D_{eval}$ is representative of the generalization capability of an individual detector to detect synthetic speech from unknown speech generators. 
As 11 out of 13 speech generators used in $D_{eval}$ set are unknown and speech signals from them were not used during training.
All methods except MFCC-ResNet have almost perfect performance on $D_{dev}$ set (\gls{eer} 0\%).
The $EER$ on $D_{eval}$ set, as expected, is higher than on $D_{dev}$ set for each method.
For both time domain and spectrogram-based approaches, the detectors which use self-supervised networks have substantially better performance on $D_{eval}$ set, however, that comes at a computational cost: large training time and substantially higher model parameters.
For 3 methods (in ~\cref{tab:results-asv2019}), our implementation has different \gls{eer} on $D_{eval}$ set than reported in their original work. 
The LFCC-GMMs implemented by us has \gls{eer} of $3.67\%$ on $D_{eval}$ set as shown in~\cref{tab:results-asv2019}, while the \gls{eer} reported in~\cite{asvspoof19} is $8.09\%$.
Implementation in~\cite{asvspoof19} was done in MATLAB, but we used Python and scikit-learn for our implementation.
Our ablation study reported in supplement material shows that the hyper-parameters are not the reason for this change as we also got better performance using same hyperparameters reported in ~\cite{asvspoof19}.
For Spec-ResNet and MFCC-ResNet, the \gls{eer} from our models weights (refer~\cref{tab:results-asv2019}) on $D_{eval}$ set are $0.42$ and $2.25$ percentage points, respectively, higher than \gls{eer} reported in ~\cite{alzantot2019}.
We use same hyper-parameters but an updated version of the feature extraction package as compared to the version used by~\cite{alzantot2019}, as discussed in the supplement material.
\subsection{Experiment 2: Studying Bias on Gender}
\begin{table}[!t]
    \centering
    \caption{Gender Bias Study for all detectors. Values are in \%.}

\resizebox{1.0\columnwidth}{!}{%
\begin{tabular}{@{\extracolsep{-4pt}}lccccccccc}
\toprule
\makecell{\bfseries DN} &
\makecell{\bfseries Metric} & 
\makecell{\bfseries \boldsymbol{$D_{US-20s-M}$}} &
\makecell{\bfseries \boldsymbol{$D_{US-20s-F}$}} & &
\makecell{\bfseries \boldsymbol{$D_{US-30s-M}$}} & 
\makecell{\bfseries \boldsymbol{$D_{US-30s-F}$}} & &
\makecell{\bfseries \boldsymbol{$D_{US-60s-M}$}} & 
\makecell{\bfseries \boldsymbol{$D_{US-60s-F}$}} 
 \\ 
 \midrule
 \multirow{4}{*}{{D01}} & $\Delta FPR_{1}$ & $1.47 \pm 0.087$ & $0.00 \pm 0.012$ & & $0.92 \pm 0.034$ & $0.00 \pm 0.018$ & & \boldsymbol{$6.56 \pm 0.076$} & \boldsymbol{$0.00 \pm 0.105$}    \\
 & $\Delta FPR_{2}$ & \boldsymbol{$0.36 \pm 0.009$} & \boldsymbol{$0.00 \pm 0.006$} & & $0.10 \pm 0.012$ & $0.00 \pm 0.005$ & & $0.00 \pm 0.006$ & $0.03 \pm 0.009$ \\
 & $\Delta FPR_{3}$ & $0.80 \pm 0.162$ & $0.00 \pm 0.053$ & & $6.01 \pm 0.282$ & $0.00 \pm 0.101$ & & \boldsymbol{$26.12 \pm 0.248$} & \boldsymbol{$0.00 \pm 0.348$}    \\
 & $\Delta EER$ & $1.45 \pm 0.053$ & $0.00 \pm 0.026$ & & $0.00 \pm 0.171$ & $1.21 \pm 0.024$ & & \boldsymbol{$14.84 \pm 0.062$} & \boldsymbol{$0.00 \pm 0.083$} \\

 \midrule
 \multirow{4}{*}{{D02}} & $\Delta FPR_{1}$ & $0.00 \pm 0.057$ & $2.09 \pm 0.047$ & & $0.00 \pm 0.313$ & $2.40 \pm 0.051$ & & \boldsymbol{$11.40 \pm 0.208$} & \boldsymbol{$0.00 \pm 0.288$}    \\
 & $\Delta FPR_{2}$ & $0.00 \pm 0.132$ & $0.61 \pm 0.026$ & & \boldsymbol{$0.00 \pm 0.381$} & \boldsymbol{$6.73 \pm 0.048$} & & {$0.38 \pm 0.101$} & $0.00 \pm 0.136$ \\
 & $\Delta FPR_{3}$ & $0.32 \pm 0.112$ & $0.00 \pm 0.020$ & &\boldsymbol{$0.93 \pm 0.095$} & \boldsymbol{$0.00 \pm 0.041$} & & $0.78 \pm 0.019$ & $0.00 \pm 0.013$   \\
 & $\Delta EER$ & $0.20 \pm 0.039$ & $0.00 \pm 0.013$ & & $0.71 \pm 0.080$ & $0.00 \pm 0.015$ & & \boldsymbol{$1.16 \pm 0.038$} & \boldsymbol{$0.00 \pm 0.050$}    \\
 
 \midrule
 \multirow{4}{*}{{D03}} & $\Delta FPR_{1}$ & \boldsymbol{$0.15 \pm 0.030$} & \boldsymbol{$0.00 \pm 0.005$} & & $0.05 \pm 0.009$ & $0.00 \pm 0.005$ & & $0.12 \pm 0.010$ & $0.00 \pm 0.013$   \\
 & $\Delta FPR_{2}$ & \boldsymbol{$0.14 \pm 0.022$} & \boldsymbol{$0.00 \pm 0.009$} & & $0.11 \pm 0.059$ & $0.00 \pm 0.008$ & & $0.13 \pm 0.018$ & $0.00 \pm 0.025$ \\
 & $\Delta FPR_{3}$ & \boldsymbol{$0.08 \pm 0.016$} & \boldsymbol{$0.00 \pm 0.005$} & & $0.02 \pm 0.012$ & $0.00 \pm 0.008$ & & $0.08 \pm 0.017$ & $0.00 \pm 0.025$    \\
 & $\Delta EER$ & \boldsymbol{$2.96 \pm 0.035$} & \boldsymbol{$0.00 \pm 0.012$} & & $1.02 \pm 0.053$ & $0.00 \pm 0.006$ & & $2.02 \pm 0.020$ & $0.00 \pm 0.026$   \\
 \midrule
 
 \multirow{4}{*}{{D04}} & $\Delta FPR_{1}$ & $22.48 \pm 0.001$ & $0.00 \pm 0.034$ & & \boldsymbol{$39.87 \pm 0.280$} & \boldsymbol{$0.00 \pm 0.032$} & & $11.61 \pm 0.135$ & $0.00 \pm 0.177$    \\
 & $\Delta FPR_{2}$ & $22.32 \pm 0.195$ & $0.00 \pm 0.052$ & & \boldsymbol{$39.52 \pm 0.253$} & \boldsymbol{$0.00 \pm 0.072$} & & $11.17 \pm 0.157$ & $0.00 \pm 0.207$ \\
 & $\Delta FPR_{3}$ & $22.80 \pm 0.150$ & $0.00 \pm 0.023$ & & \boldsymbol{$40.34 \pm 0.246$} & \boldsymbol{$0.00 \pm 0.071$} & & $11.78 \pm 0.173$ & $0.00 \pm 0.229$ \\
 & $\Delta EER$ & $4.94 \pm 0.089$ & $0.00 \pm 0.028$ & & \boldsymbol{$14.80 \pm 0.238$} & \boldsymbol{$0.00 \pm 0.035$} & & $0.97 \pm 0.049$ & $0.00 \pm 0.057$    \\
 \midrule

  \multirow{4}{*}{{D05}} & $\Delta FPR_{1}$ & $0.00 \pm 0.000$ & $0.00 \pm 0.000$ & & $0.00 \pm 0.000$ & $0.00 \pm 0.000$ & & $0.00 \pm 0.000$ & $0.00 \pm 0.000$   \\
 & $\Delta FPR_{2}$ & $0.00 \pm 0.000$ & $0.00 \pm 0.000$ & & $0.00 \pm 0.000$ & $0.00 \pm 0.000$ & & $0.00 \pm 0.000$ & $0.00 \pm 0.000$ \\
 & $\Delta FPR_{3}$ & $0.00 \pm 0.000$ & $0.00 \pm 0.000$ & & $0.00 \pm 0.000$ & $0.00 \pm 0.000$ & & $0.00 \pm 0.000$ & $0.00 \pm 0.000$    \\
 & $\Delta EER$ & $2.04 \pm 0.072$ & $0.00 \pm 0.008$ & & $0.70 \pm 0.115$ & $0.00 \pm 0.031$ & & \boldsymbol{$0.00 \pm 0.042$} & \boldsymbol{$2.56 \pm 0.049$}   \\
 \midrule

 \multirow{4}{*}{{D06}} & $\Delta FPR_{1}$ & $0.00 \pm 0.099$ & $0.41 \pm 0.028$ & & $7.39 \pm 0.178$ & $0.00 \pm 0.092$ & & \boldsymbol{$0.00 \pm 0.178$} & \boldsymbol{$8.58 \pm 0.190$}  \\
 & $\Delta FPR_{2}$ & $0.12 \pm 0.171$ & $0.00 \pm 0.035$ & & \boldsymbol{$10.74 \pm 0.238$} & \boldsymbol{$0.00 \pm 0.029$} & & $0.00 \pm 0.158$ & $5.74 \pm 0.205$ \\
 & $\Delta FPR_{3}$ & $0.00 \pm 0.102$ & $0.90 \pm 0.015$ & & $4.91 \pm 0.170$ & $0.00 \pm 0.055$ & & \boldsymbol{$0.00 \pm 0.109$} & \boldsymbol{$9.04 \pm 0.147$} \\
 & $\Delta EER$ & $1.39 \pm 0.113$ & $0.00 \pm 0.028$ & & \boldsymbol{$6.62 \pm 0.139$} & \boldsymbol{$0.00 \pm 0.046$} & & $1.10 \pm 0.092$ & $0.00 \pm 0.097$   \\

\bottomrule
\end{tabular}
}
    \label{tab:gender_bias}
\end{table}
In this experiment, we determine if a detector misclassifies bona fide speech more from one particular gender.
The results for this experiment are shown in~\cref{tab:gender_bias}.

We fixed the accent to US English and studied gender bias on speaker in three different age groups, namely, speakers in their 20s, 30s, and 60s.
For each age group, we have two sets. 
For example, $D_{US-20s-M}$ and $D_{US-20s-F}$ are two sets for speakers in age group 20s.
We obtain $\Delta EER$, $\Delta FPR_1$, $\Delta FPR_2$, and $\Delta FPR_3$ as a measure of bias in different scenarios.
For example, $FPR$s are obtained using threshold from an independent dataset.
However, \gls{eer} assumes that ground truth labels are known and it is the rate where \gls{fpr} and \gls{fnr} are equal (see~\cref{sec:metrics}).

In~\cref{tab:gender_bias}, we highlight most gender biased age group for each detector.
For example, Detector Number (DN) D01 \ie, ~\gls{tssdnet} has 26.12 percentage points higher $FPR_3$ for male speakers than for female speakers in 60s age group.
Similar bias is also evident in $\Delta EER$ value of this group.
Similar results are also obtained for male and female speakers in age group 20s, where D01 has 1.45 percentage points higher \gls{eer} on male speakers than on female speakers and all $FPR$s are higher on male speakers.
For speakers in $30s$, the $FPR$ is similarly higher for male speakers.
Overall, these observations suggest that D01 misclassifies bona fide speech from male speakers more than that from female speakers.
Similar observations can also be noted for D04 \ie, PS3DT which has approximately 50 percentage points of more misclassifications for bona fide speech from male speakers than from female speakers.
D03, \ie, Spec-ResNet also has higher misclassification rate for male speakers in 20s, 30s and 60s than female speakers. However, this bias is not as high as other detectors.
One interesting point to note here is that if for a detector the $FPR$ is $100\%$ for both bona fide speech from male and female speaker, then all $\Delta FPR$ metrics will be zero. 
This does not imply that the detector is not biased, but in such cases $\Delta EER$ will be able to show bias. 
This behaviour is observed for detector D05 \ie, LFCC-GMMs, stressing on the fact that simple \gls{gmms} based synthetic speech detectors may not be suitable for deployment for in-the-wild synthetic speech detection scenarios (also observed in~\cite{asvspoof_2021}).
Overall, it can be observed from~\cref{tab:gender_bias} that majority of the approaches misclassify bona fide speech from male speakers more than they do for female speakers.
Coincidently, similar bias for male gender was also found in deepfake face detectors~\cite{deepfake_fair_23}.
\begin{table}[!t]
    \centering
    \caption{Age Bias Study for Male Speakers. Values are in \%.}

\resizebox{\columnwidth}{!}{%
\begin{tabular}{@{\extracolsep{-4pt}}lccccccc}
\toprule
\makecell{\bfseries DN} &
\makecell{\bfseries Metric} & 
\makecell{\boldsymbol{$D_{US-ts-M}$}} & 
\makecell{\bfseries \boldsymbol{$D_{US-20s-M}$}} & 
\makecell{\bfseries \boldsymbol{$D_{US-30s-M}$}} & 
\makecell{\bfseries \boldsymbol{$D_{US-40s-M}$}} &
\makecell{\bfseries \boldsymbol{$D_{US-50s-M}$}} & 
\makecell{\bfseries \boldsymbol{$D_{US-60s-M}$}}  
 \\ 
 
 \midrule
 \multirow{4}{*}{D01} & $\Delta FPR_{1}$ & {$0.00 \pm 0.109$} & $1.03 \pm 0.093$ & $1.32 \pm 0.179$ & $0.54 \pm 0.092$ & $0.54 \pm 0.092$ & \boldsymbol{$1.36 \pm 0.131$} \\
 & $\Delta FPR_{2}$ & $0.03 \pm 0.026$ & $0.04 \pm 0.030$ & \boldsymbol{$0.07 \pm 0.017$} & $0.03 \pm 0.019$ & $0.01 \pm 0.020$ & $0.00 \pm 0.021$ \\
 & $\Delta FPR_{3}$ & $2.51 \pm 0.261$ & $6.11 \pm 0.438$ & $6.62 \pm 0.375$ & $0.00 \pm 0.282$ & $2.21 \pm 0.231$ & \boldsymbol{$15.34 \pm 0.361$}    \\
 & $\Delta EER$ & $2.57 \pm 0.368$ & $4.50 \pm 0.473$ & $1.22 \pm 0.368$ & $0.00 \pm 0.475$ & $2.37 \pm 0.359$ & \boldsymbol{$15.89 \pm 0.360$}    \\
 
 \midrule
 \multirow{4}{*}{D02} & $\Delta FPR_{1}$ & \boldsymbol{$11.88 \pm 0.395$} & $5.30 \pm 0.323$ & $0.00 \pm 0.249$ & $4.32 \pm 0.354$ & $5.31 \pm 0.278$ & $6.90 \pm 0.384$    \\
 & $\Delta FPR_{2}$ & $8.18 \pm 0.463$ & $6.80 \pm 0.457$ & $0.00 \pm 0.455$ & $1.23 \pm 0.440$ & $3.54 \pm 0.370$ & \boldsymbol{$9.25 \pm 0.336$} \\
 & $\Delta FPR_{3}$ & \boldsymbol{$2.62 \pm 0.171$} & $1.16 \pm 0.147$ & $0.88 \pm 0.198$ & $2.56 \pm 0.196$ & $1.76 \pm 0.152$ & $0.00 \pm 0.103$   \\
 & $\Delta EER$ & \boldsymbol{$2.10 \pm 0.106$} & $0.83 \pm 0.129$ & $0.71 \pm 0.154$ & $1.93 \pm 0.190$ & $1.40 \pm 0.109$ & $0.00 \pm 0.136$    \\
 \midrule
 \multirow{4}{*}{D03} & $\Delta FPR_{1}$ & \boldsymbol{$0.45 \pm 0.062$} & $0.30 \pm 0.088$ & $0.29 \pm 0.075$ & $0.00 \pm 0.085$ & $0.36 \pm 0.065$ & $0.43 \pm 0.066$   \\
 & $\Delta FPR_{2}$ & $0.55 \pm 0.068$ & $0.36 \pm 0.107$ & $0.34 \pm 0.086$ & $0.00 \pm 0.087$ & $0.42 \pm 0.069$ & \boldsymbol{$0.57 \pm 0.066$} \\
 & $\Delta FPR_{3}$ & \boldsymbol{$0.22 \pm 0.049$} & $0.12 \pm 0.051$ & $0.10 \pm 0.057$ & $0.00 \pm 0.060$ & $0.16 \pm 0.043$ & $0.22 \pm 0.046$    \\
 & $\Delta EER$ & $2.27 \pm 0.160$ & $2.02 \pm 0.141$ & $0.34 \pm 0.178$ & $0.00 \pm 0.172$ & $0.98 \pm 0.132$ & \boldsymbol{$3.19 \pm 0.132$}   \\
 
 \midrule
  \multirow{4}{*}{D04} & $\Delta FPR_{1}$ & $0.00 \pm 0.773$ & $5.52 \pm 0.776$ & \boldsymbol{$12.11 \pm 0.622$} & $10.24 \pm 0.736$ & $11.21 \pm 0.596$ & $6.70 \pm 0.576$   \\
 & $\Delta FPR_{2}$ & $0.00 \pm 0.599$ & $5.58 \pm 0.588$ & $11.87 \pm 0.548$ & $10.49 \pm 0.559$ & \boldsymbol{$11.42 \pm 0.432$} & $7.15 \pm 0.621$ \\
 & $\Delta FPR_{3}$ & $0.00 \pm 0.413$ & $5.28 \pm 0.841$ & \boldsymbol{$12.34 \pm 0.416$} & $10.99 \pm 0.356$ & $11.16 \pm 0.338$ & $6.91 \pm 0.355$    \\
 & $\Delta EER$ & $0.24 \pm 0.188$ & $1.26 \pm 0.132$ & \boldsymbol{$7.14 \pm 0.200$} & $1.43 \pm 0.253$ & $3.04 \pm 0.110$ & $0.00 \pm 0.103$   \\
 
 \midrule
 \multirow{4}{*}{D05} & $\Delta FPR_{1}$ & $0.00 \pm 0.000$ & $0.00 \pm 0.000$ & $0.00 \pm 0.000$ & $0.00 \pm 0.000$ & $0.00 \pm 0.000$ & $0.00 \pm 0.000$    \\
 & $\Delta FPR_{2}$ & $0.00 \pm 0.000$ & $0.00 \pm 0.000$ &  $0.00 \pm 0.000$ & $0.00 \pm 0.000$ &  $0.00 \pm 0.000$ & $0.00 \pm 0.000$ \\
 & $\Delta FPR_{3}$ & $0.00 \pm 0.000$ & $0.00 \pm 0.000$ &  $0.00 \pm 0.000$ & $0.00 \pm 0.000$ &  $0.00 \pm 0.000$ & $0.00 \pm 0.000$ \\
 & $\Delta EER$ & $0.00 \pm 0.165$ & $1.43 \pm 0.213$ &  \boldsymbol{$3.19 \pm 0.131$} & $1.07 \pm 0.169$ &  $1.89 \pm 0.138$ & $3.07 \pm 0.165$    \\
 \midrule
 
 \multirow{4}{*}{D06} & $\Delta FPR_{1}$ & {$15.27 \pm 0.415$} & $13.29 \pm 0.434$ &  \boldsymbol{$17.97 \pm 0.509$} & $12.39 \pm 0.406$ &  $11.41 \pm 0.338$ & $0.00 \pm 0.446$  \\
 & $\Delta FPR_{2}$ & $16.23 \pm 0.455$ & $14.48 \pm 0.472$ &  \boldsymbol{$19.70 \pm 0.402$} & $13.13 \pm 0.684$ & $11.30 \pm 0.425$ & $0.00 \pm 0.441$ \\
 & $\Delta FPR_{3}$ & $12.22 \pm 0.327$ & $10.64 \pm 0.323$ &  \boldsymbol{$13.71 \pm 0.286$} & $9.89 \pm 0.425$ & $9.05 \pm 0.293$ & $0.00 \pm 0.328$ \\
 & $\Delta EER$ & $8.01 \pm 0.185$ & $7.58 \pm 0.297$ & \boldsymbol{$10.33 \pm 0.162$} & $7.33 \pm 0.218$ & $4.59 \pm 0.223$ & $0.00 \pm 0.196$   \\
\bottomrule
\end{tabular}
}
    \label{tab:age_m_bias}
\end{table}
\subsection{Experiment 3: Studying Bias on Age}
In this experiment, we determine if a detector misclassifies bona fide speech from one particular age group more than other age groups.
We fixed the accent to US English and studied age bias on two genders male and female.
For each gender, we have six sets having bona fide speech from speakers in teens, 20s, 30s, 40s, 50s, and 60s.

\cref{tab:age_m_bias} and~\cref{tab:age_f_bias} show results from our age bias study on male and female gender, respectively.
We observe that majority of the detectors have more misclassification for speakers in extreme age groups \ie, speakers who are either teens or in 60s.
In contrast to this, detector D04 misclassifies bona fide speech from speakers in their 30s more than speakers in other age groups.
This behaviour is only observed for male speakers (\cref{tab:age_m_bias}) and not for female speakers (\cref{tab:age_f_bias}).
For both male and female groups, detector D03~\cite{alzantot2019} has least bias among different age groups.
Detector D05, \ie, LFCC-GMMs again has $100\%$ \gls{fpr} on all accents, resulting in $\Delta FPR$ of zero. However, $\Delta EER$ shows that this method misclassifies more bona fide speech for speakers in 60s and 30s age groups.
This is true for both male and female speakers.
Overall, majority of detectors unfairly misclassify bona fide speech from speakers in teens and 60s age group.
Depending on detector and bias measure scenario, the bias varies from minimum $0.3$ percentage points (D01 in~\cref{tab:age_f_bias}) to approximately $23$ percentage points (D04 in~\cref{tab:accent_f_bias}).

\begin{table}[!t]
    \centering
    \caption{Age Bias Study for Female Speakers. Values are in \%.}

\resizebox{\columnwidth}{!}{%
\begin{tabular}{@{\extracolsep{-4pt}}lccccccc}
\toprule
\makecell{\bfseries DN} &
\makecell{\bfseries Metric} & 
\makecell{\boldsymbol{$D_{US-ts-F}$}} & 
\makecell{\bfseries \boldsymbol{$D_{US-20s-F}$}} & 
\makecell{\bfseries \boldsymbol{$D_{US-30s-F}$}} & 
\makecell{\bfseries \boldsymbol{$D_{US-40s-F}$}} &
\makecell{\bfseries \boldsymbol{$D_{US-50s-F}$}} & 
\makecell{\bfseries \boldsymbol{$D_{US-60s-F}$}}  
 \\  
 
 \midrule
 \multirow{4}{*}{D01} & $\Delta FPR_{1}$ & \boldsymbol{$6.54 \pm 0.356$} & $4.88 \pm 0.421$ & $5.78 \pm 0.350$ & $5.75 \pm 0.346$ & $3.97 \pm 0.351$ & $0.00 \pm 0.488$ \\
 & $\Delta FPR_{2}$ & {$0.30 \pm 0.075$} & $0.00 \pm 0.105$ & $0.27 \pm 0.076$ & $0.19 \pm 0.074$ & $0.22 \pm 0.076$ & \boldsymbol{$0.32 \pm 0.076$} \\
 & $\Delta FPR_{3}$ & \boldsymbol{$19.83 \pm 0.238$} & $18.58 \pm 0.340$ & $14.16 \pm 0.252$ & $0.00 \pm 0.086$ & $10.51 \pm 0.304$ & $2.56 \pm 0.252$    \\
 & $\Delta EER$ & $5.92 \pm 0.230$ & \boldsymbol{$6.82 \pm 0.215$} & $6.09 \pm 0.124$ & $0.00 \pm 0.049$ & $4.40 \pm 0.126$ & $4.73 \pm 0.077$    \\
 
 \midrule
 \multirow{4}{*}{D02} & $\Delta FPR_{1}$ & \boldsymbol{$18.48 \pm 0.459$} & $12.16 \pm 0.538$ & $6.42 \pm 0.431$ & $13.06 \pm 0.325$ & $5.87 \pm 0.389$ & $0.00 \pm 0.458$    \\
 & $\Delta FPR_{2}$ & {$3.71 \pm 0.286$} & $2.77 \pm 0.258$ & $1.29 \pm 0.347$ & $5.41 \pm 0.225$ & $0.00 \pm 0.317$ & \boldsymbol{$4.04 \pm 0.305$} \\
 & $\Delta FPR_{3}$ & \boldsymbol{$12.15 \pm 0.233$} & $1.71 \pm 0.185$ & $0.75 \pm 0.078$ & $1.24 \pm 0.044$ & $2.07 \pm 0.124$ & $0.00 \pm 0.058$   \\
 & $\Delta EER$ & \boldsymbol{$9.61 \pm 0.168$} & $1.70 \pm 0.128$ & $0.96 \pm 0.138$ & $1.28 \pm 0.074$ & $2.05 \pm 0.094$ & $0.00 \pm 0.104$    \\
 \midrule
 \multirow{4}{*}{D03} & $\Delta FPR_{1}$ & \boldsymbol{$0.80 \pm 0.073$} & $0.63 \pm 0.110$ & $0.67 \pm 0.075$ & $0.69 \pm 0.063$ & $0.00 \pm 0.089$ & $0.78 \pm 0.077$   \\
 & $\Delta FPR_{2}$ & $0.91 \pm 0.056$ & $0.73 \pm 0.046$ & $0.70 \pm 0.040$ & $0.82 \pm 0.031$ & $0.00 \pm 0.044$ & \boldsymbol{$0.96 \pm 0.062$} \\
 & $\Delta FPR_{3}$ & \boldsymbol{$0.35 \pm 0.072$} & $0.26 \pm 0.077$ & $0.27 \pm 0.076$ & $0.30 \pm 0.071$ & $0.00 \pm 0.100$ & $0.33 \pm 0.076$    \\
 & $\Delta EER$ & $0.00 \pm 0.097$ & $1.23 \pm 0.093$ & $1.58 \pm 0.120$ & $2.61 \pm 0.071$ & $0.21 \pm 0.158$ & \boldsymbol{$3.42 \pm 0.102$}   \\
 
 \midrule
  \multirow{4}{*}{D04} & $\Delta FPR_{1}$ & $20.47 \pm 0.377$ & $11.28 \pm 0.432$ & $0.00 \pm 0.441$ & $13.79 \pm 0.312$ & $1.22 \pm 0.358$ & \boldsymbol{$22.88 \pm 0.594$}   \\
 & $\Delta FPR_{2}$ & $20.29 \pm 0.396$ & $11.31 \pm 0.354$ & $0.00 \pm 0.362$ & $14.48 \pm 0.262$ & $1.88 \pm 0.491$ & \boldsymbol{$23.65 \pm 0.309$} \\
 & $\Delta FPR_{3}$ & $20.79 \pm 0.514$ & $11.08 \pm 0.642$ & $0.00 \pm 0.466$ & $13.44 \pm 0.334$ & $1.21 \pm 0.674$ & \boldsymbol{$23.15 \pm 0.436$}    \\
 & $\Delta EER$ & \boldsymbol{$12.49 \pm 0.179$} & $4.88 \pm 0.130$ & $1.09 \pm 0.173$ & $2.58 \pm 0.031$ & $0.00 \pm 0.039$ & $7.66 \pm 0.142$   \\
 
 \midrule
 \multirow{4}{*}{D05} & $\Delta FPR_{1}$ & $0.00 \pm 0.000$ & $0.00 \pm 0.000$ & $0.00 \pm 0.000$ & $0.00 \pm 0.000$ & $0.00 \pm 0.000$ & $0.00 \pm 0.000$    \\
 & $\Delta FPR_{2}$ & $0.00 \pm 0.000$ & $0.00 \pm 0.000$ &  $0.00 \pm 0.000$ & $0.00 \pm 0.000$ &  $0.00 \pm 0.000$ & $0.00 \pm 0.000$ \\
 & $\Delta FPR_{3}$ & $0.00 \pm 0.000$ & $0.00 \pm 0.000$ &  $0.00 \pm 0.000$ & $0.00 \pm 0.000$ &  $0.00 \pm 0.000$ & $0.00 \pm 0.000$ \\
 & $\Delta EER$ & $0.00 \pm 0.082$ & $0.58 \pm 0.093$ &  $3.68 \pm 0.152$ & $1.98 \pm 0.059$ &  $2.38 \pm 0.119$ & \boldsymbol{$6.75 \pm 0.131$}    \\
 \midrule
 
 \multirow{4}{*}{D06} & $\Delta FPR_{1}$ & \boldsymbol{$11.36 \pm 0.366$} & $8.00 \pm 0.412$ &  $4.12 \pm 0.454$ & $2.60 \pm 0.355$ &  $0.00 \pm 0.502$ & $2.74 \pm 0.504$  \\
 & $\Delta FPR_{2}$ & \boldsymbol{$14.21 \pm 0.405$} & $10.00 \pm 0.436$ &  $4.89 \pm 0.408$ & $3.06 \pm 0.310$ & $0.00 \pm 0.433$ & $0.82 \pm 0.466$ \\
 & $\Delta FPR_{3}$ & \boldsymbol{$7.98 \pm 0.120$} & $6.21 \pm 0.263$ &  $3.55 \pm 0.214$ & $2.47 \pm 0.095$ & $0.00 \pm 0.125$ & $3.63 \pm 0.223$ \\
 & $\Delta EER$ & \boldsymbol{$9.86 \pm 0.291$} & $7.28 \pm 0.334$ & $4.94 \pm 0.293$ & $4.08 \pm 0.273$ & $2.45 \pm 0.319$ & $0.00 \pm 0.385$   \\
\bottomrule
\end{tabular}
}
    \label{tab:age_f_bias}
\end{table}
 
\subsection{Experiment 4: Studying Bias on Accent}
In this experiment, we determine if a detector unfairly targets English bona fide speech with one particular accent.

We fixed the age group to speakers in their 20s, and performed an independent accent bias study on male and female genders.
For each gender, we performed an accent bias study on five different accents: Canadian (CN), United States (US), British (UK), Australian (AU), and South Asian (SA).

\cref{tab:accent_m_bias} and ~\cref{tab:accent_f_bias} show results from our accent bias study on male and female gender, respectively.
The results show that the detectors are most accurate for US accent English.
This may be because most of the open source datasets such as ASVspoof2019~\cite{asvdata_2019} and LibriSpeech~\cite{librispeech} on which these detectors are fine-tuned or pretrained have majority speech samples 
with US accent.
Almost all detectors have higher \gls{fpr} for bona fide speech from South Asian (SA) and Australian (AU) English speakers. 
Also notice the \textbf{bold} values in ~\cref{tab:accent_m_bias} and ~\cref{tab:accent_f_bias}. 
Detector D02 (Wav2Vec2-AASIST~\cite{tak22_odyssey}) misclassifies bona fide speech from South Asian female speakers approximately 52 percentage points higher than Canadian female speakers. 
This observation is also true for bona fide speech from South Asian male speakers with approximately 33 percentage points higher misclassification than Australian male speakers.
Apart from high \gls{fpr}s for bona fide speech from Australian and South Asian speakers, detector $D04$, (PS3DT)~\cite{ps3dt} also has around 11 percentage points higher misclassification for bona fide speech with British accents than Canadian accent~\cref{tab:accent_m_bias}.
Overall, the majority of detectors unfairly target bona fide speech from speakers with South Asian and Australian accents.
\begin{table}[!t]
    \centering
    \caption{Accent Bias Study for Male Speakers. Values are in \%.}
\resizebox{\columnwidth}{!}{%
\begin{tabular}{@{\extracolsep{-4pt}}lcccccc}
\toprule
\makecell{\bfseries DN} &
\makecell{\bfseries Metric} & 
\makecell{\bfseries \boldsymbol{$D_{CN-20s-M}$}} & 
\makecell{\bfseries \boldsymbol{$D_{US-20s-M}$}} & 
\makecell{\bfseries \boldsymbol{$D_{UK-20s-M}$}} & 
\makecell{\bfseries \boldsymbol{$D_{AU-20s-M}$}} &
\makecell{\bfseries \boldsymbol{$D_{SA-20s-M}$}}   
 \\ 
 \midrule
 \multirow{4}{*}{{D01}} & $\Delta FPR_{1}$ & $0.71 \pm 0.070$ & $0.40 \pm 0.180$ & $0.29 \pm 0.177$ & $0.00 \pm 0.020$ & \boldsymbol{$0.74 \pm 0.106$} \\
 & $\Delta FPR_{2}$ & $0.05 \pm 0.012$ & $0.00 \pm 0.015$ &  $0.02 \pm 0.021$ & \boldsymbol{$0.06 \pm 0.010$} &  $0.02 \pm 0.015$ \\
 & $\Delta FPR_{3}$ & $6.16 \pm 0.379$ & $4.76 \pm 0.699$ &  $0.00 \pm 0.497$ & $2.07 \pm 0.353$ &  \boldsymbol{$9.63 \pm 0.455$} \\
 & $\Delta EER$ & $5.21 \pm 0.227$ & $3.62 \pm 0.335$ &  $0.00 \pm 0.273$ & $5.40 \pm 0.195$ &  \boldsymbol{$9.17 \pm 0.343$} \\
 
 \midrule
 \multirow{4}{*}{{D02}} & $\Delta FPR_{1}$ & $5.65 \pm 0.214$ & $5.33 \pm 0.451$ & $5.09 \pm 0.233$ & $0.00 \pm 0.045$ & \boldsymbol{$33.31 \pm 0.456$} \\
 & $\Delta FPR_{2}$ & $2.90 \pm 0.209$ & $2.55 \pm 0.334$ &  $2.78 \pm 0.151$ & $0.00 \pm 0.038$ &  \boldsymbol{$7.31 \pm 0.120$} \\
 & $\Delta FPR_{3}$ & $0.95 \pm 0.046$ & $1.74 \pm 0.236$ &  $1.79 \pm 0.224$ & $0.00 \pm 0.010$ &  \boldsymbol{$5.86 \pm 0.108$} \\
 & $\Delta EER$ & $1.06 \pm 0.026$ & $1.23 \pm 0.157$ &  $1.50 \pm 0.174$ & $0.00 \pm 0.018$ &  \boldsymbol{$4.80 \pm 0.112$} \\
 \midrule
 
 \multirow{4}{*}{{D03}} & $\Delta FPR_{1}$ & $0.13 \pm 0.076$ & $0.00 \pm 0.107$ & $0.11 \pm 0.088$ & \boldsymbol{$0.14 \pm 0.075$} & $0.02 \pm 0.095$ \\
 & $\Delta FPR_{2}$ & $0.13 \pm 0.076$ & $0.01 \pm 0.097$ &  {$0.16 \pm 0.092$} & \boldsymbol{$0.18 \pm 0.075$} &  $0.00 \pm 0.106$ \\
 & $\Delta FPR_{3}$ & \boldsymbol{$0.06 \pm 0.026$} & $0.00 \pm 0.036$ &  $0.05 \pm 0.041$ & $0.04 \pm 0.026$ &  $0.00 \pm 0.055$ \\
 & $\Delta EER$ & \boldsymbol{$1.78 \pm 0.097$} & $0.96 \pm 0.111$ &  $1.26 \pm 0.110$ & $1.66 \pm 0.091$ &  $0.00 \pm 0.128$ \\
 
 \midrule
  \multirow{4}{*}{{D04}} & $\Delta FPR_{1}$ & $0.00 \pm 0.108$ & $3.34 \pm 0.792$ & \boldsymbol{$11.51 \pm 0.214$} & $9.98 \pm 0.087$ & $6.09 \pm 0.289$ \\
 & $\Delta FPR_{2}$ & $0.00 \pm 0.256$ & $2.29 \pm 0.315$ &  \boldsymbol{$11.14 \pm 0.252$} & $9.71 \pm 0.185$ &  $5.50 \pm 0.439$ \\
 & $\Delta FPR_{3}$ & $0.00 \pm 0.355$ & $3.32 \pm 0.627$ &  \boldsymbol{$11.74 \pm 0.338$} & $10.29 \pm 0.257$ & {$5.68 \pm 0.310$} \\
 & $\Delta EER$ & $0.00 \pm 0.258$ & $0.39 \pm 0.336$ &  $1.40 \pm 0.300$ & \boldsymbol{$2.98 \pm 0.183$} &  {$2.49 \pm 0.248$} \\
 
 \midrule
 \multirow{4}{*}{{D05}} & $\Delta FPR_{1}$ & $0.00 \pm 0.000$ & $0.00 \pm 0.000$ & $0.00 \pm 0.000$ & $0.00 \pm 0.000$ & $0.00 \pm 0.000$ \\
 & $\Delta FPR_{2}$ & $0.00 \pm 0.000$ & $0.00 \pm 0.000$ &  $0.00 \pm 0.000$ & $0.00 \pm 0.000$ &  $0.00 \pm 0.000$ \\
 & $\Delta FPR_{3}$ & $0.00 \pm 0.007$ & $0.00 \pm 0.007$ &  $0.00 \pm 0.007$ & $0.00 \pm 0.007$ &  $0.00 \pm 0.010$ \\
 & $\Delta EER$ & $5.05 \pm 0.137$ & $5.21 \pm 0.256$ &  $4.27 \pm 0.180$ & \boldsymbol{$6.48 \pm 0.125$} &  $0.00 \pm 0.176$ \\
 \midrule
 
 \multirow{4}{*}{{D06}} & $\Delta FPR_{1}$ & $2.26 \pm 0.189$ & $0.36 \pm 0.333$ & $5.21 \pm 0.289$ & $0.00 \pm 0.056$ & \boldsymbol{$5.30 \pm 0.215$} \\
 & $\Delta FPR_{2}$ & $2.74 \pm 0.275$ & $0.91 \pm 0.487$ &  \boldsymbol{$6.38 \pm 0.299$} & $0.00 \pm 0.072$ &  $6.23 \pm 0.291$ \\
 & $\Delta FPR_{3}$ & $2.09 \pm 0.535$ & $0.00 \pm 0.731$ &  \boldsymbol{$4.26 \pm 0.559$} & $0.48 \pm 0.517$ &  $4.17 \pm 0.553$ \\
 & $\Delta EER$ & $2.07 \pm 0.118$ & $0.69 \pm 0.119$ &  $3.15 \pm 0.107$ & $0.00 \pm 0.032$ &  \boldsymbol{$3.09 \pm 0.155$} \\
\bottomrule
\end{tabular}
}
    \label{tab:accent_m_bias}
\end{table}

\begin{table}[!t]
    \centering
    \caption{Accent Bias Study for Female Speakers. Values are in \%.}

\resizebox{\columnwidth}{!}{%
\begin{tabular}{@{\extracolsep{-4pt}}lcccccc}
\toprule
\makecell{\bfseries DN} &
\makecell{\bfseries Metric} & 
\makecell{\bfseries \boldsymbol{$D_{CN-20s-M}$}} &
\makecell{\bfseries \boldsymbol{$D_{US-20s-M}$}} & 
\makecell{\bfseries \boldsymbol{$D_{UK-20s-M}$}} & 
\makecell{\bfseries \boldsymbol{$D_{AU-20s-M}$}} &
\makecell{\bfseries \boldsymbol{$D_{SA-20s-M}$}}   
 \\ 
 \midrule
 \multirow{4}{*}{D01} & $\Delta FPR_{1}$ & $0.00 \pm 0.048$ & $1.68 \pm 0.160$ & $2.79 \pm 0.071$ & \boldsymbol{$4.16 \pm 0.060$} & $3.97 \pm 0.095$ \\
 & $\Delta FPR_{2}$ & $0.00 \pm 0.013$ & $0.33 \pm 0.038$ &  $0.65 \pm 0.009$ & \boldsymbol{$0.73 \pm 0.013$} &  $0.64 \pm 0.056$ \\
 & $\Delta FPR_{3}$ & $2.47 \pm 0.228$ & $3.97 \pm 0.375$ &  $0.00 \pm 0.304$ & \boldsymbol{$14.03 \pm 0.240$} &  $13.60 \pm 0.358$ \\
 & $\Delta EER$ & $1.95 \pm 0.068$ & $2.07 \pm 0.290$ &  $0.00 \pm 0.038$ & \boldsymbol{$17.61 \pm 0.138$} &  $6.02 \pm 0.183$ \\
 
 \midrule
 \multirow{4}{*}{D02} & $\Delta FPR_{1}$ & $0.00 \pm 0.070$ & $7.78 \pm 0.422$ & $12.39 \pm 0.075$ & $40.00 \pm 0.207$ & \boldsymbol{$52.26 \pm 0.595$} \\
 & $\Delta FPR_{2}$ & $0.00 \pm 0.105$ & $1.85 \pm 0.184$ &  $1.99 \pm 0.104$ & $7.51 \pm 0.101$ &  \boldsymbol{$8.99 \pm 0.150$} \\
 & $\Delta FPR_{3}$ & $0.00 \pm 0.016$ & $1.40 \pm 0.102$ &  $3.77 \pm 0.065$ & $2.07 \pm 0.097$ &  \boldsymbol{$26.62 \pm 0.618$} \\
 & $\Delta EER$ & $0.00 \pm 0.000$ & $1.32 \pm 0.203$ &  $3.21 \pm 0.054$ & $2.39 \pm 0.061$ &  \boldsymbol{$16.30 \pm 0.389$} \\
 \midrule
 
 \multirow{4}{*}{D03} & $\Delta FPR_{1}$ & $0.88 \pm 0.053$ & $0.78 \pm 0.072$ & {$0.97 \pm 0.053$} & \boldsymbol{$1.06 \pm 0.056$} & $0.00 \pm 0.074$ \\
 & $\Delta FPR_{2}$ & $0.95 \pm 0.104$ & $0.92 \pm 0.145$ &  $1.03 \pm 0.103$ & \boldsymbol{$1.27 \pm 0.104$} &  $0.00 \pm 0.145$ \\
 & $\Delta FPR_{3}$ & $0.50 \pm 0.116$ & $0.46 \pm 0.126$ &  $0.48 \pm 0.116$ & $0.60 \pm 0.116$ &  $0.00 \pm 0.164$ \\
 & $\Delta EER$ & $2.35 \pm 0.299$ & $2.87 \pm 0.324$ &  $2.19 \pm 0.302$ & \boldsymbol{$6.32 \pm 0.301$} &  $0.00 \pm 0.421$ \\
 
 \midrule
  \multirow{4}{*}{D04} & $\Delta FPR_{1}$ & $0.00 \pm 0.117$ & $9.87 \pm 0.473$ & $20.24 \pm 0.226$ & $36.94 \pm 0.143$ & $25.20 \pm 0.560$ \\
 & $\Delta FPR_{2}$ & $0.00 \pm 0.080$ & $10.21 \pm 0.970$ &  $20.52 \pm 0.189$ & \boldsymbol{$37.07 \pm 0.148$} &  $25.80 \pm 0.468$ \\
 & $\Delta FPR_{3}$ & $0.00 \pm 0.088$ & $9.58 \pm 0.467$ &  $19.62 \pm 0.084$ & \boldsymbol{$36.78 \pm 0.273$} &  $24.71 \pm 0.458$ \\
 & $\Delta EER$ & $0.00 \pm 0.036$ & $2.71 \pm 0.212$ &  $7.61 \pm 0.079$ & $6.93 \pm 0.048$ &  \boldsymbol{$8.34 \pm 0.334$} \\
 
 \midrule
 \multirow{4}{*}{D05} & $\Delta FPR_{1}$ & $0.00 \pm 0.000$ & $0.00 \pm 0.000$ & $0.00 \pm 0.000$ & $0.00 \pm 0.000$ & $0.00 \pm 0.000$ \\
 & $\Delta FPR_{2}$ & $0.00 \pm 0.000$ & $0.00 \pm 0.000$ &  $0.00 \pm 0.000$ & $0.00 \pm 0.000$ &  $0.00 \pm 0.000$ \\
 & $\Delta FPR_{3}$ & $0.02 \pm 0.000$ & $0.02 \pm 0.000$ &  $0.00 \pm 0.000$ & $0.02 \pm 0.000$ &  $0.02 \pm 0.000$ \\
 & $\Delta EER$ & $10.98 \pm 0.088$ & $9.77 \pm 0.187$ &  \boldsymbol{$11.46 \pm 0.118$} & {$11.00 \pm 0.103$} &  $0.00 \pm 0.121$ \\
 \midrule
 
 \multirow{4}{*}{D06} & $\Delta FPR_{1}$ & $5.86 \pm 0.149$ & $5.22 \pm 0.498$ & $0.00 \pm 0.210$ & $10.71 \pm 0.165$ & \boldsymbol{$8.68 \pm 0.240$} \\
 & $\Delta FPR_{2}$ & $6.23 \pm 0.100$ & $5.64 \pm 0.541$ &  $0.00 \pm 0.089$ & \boldsymbol{$11.21 \pm 0.184$} &  $9.51 \pm 0.642$ \\
 & $\Delta FPR_{3}$ & $5.80 \pm 0.117$ & $4.33 \pm 0.247$ &  $0.00 \pm 0.151$ & \boldsymbol{$8.50 \pm 0.138$} &  $7.00 \pm 0.326$ \\
 & $\Delta EER$ & $1.59 \pm 0.074$ & $1.62 \pm 0.199$ &  $0.00 \pm 0.089$ & \boldsymbol{$4.70 \pm 0.152$} &  $2.69 \pm 0.276$ \\
\bottomrule
\end{tabular}
}
    \label{tab:accent_f_bias}
\end{table}
\subsection{Experiment 4: Bias on Stuttering Speech}
\begin{table}[!t]
    \centering
    \caption{Stuttering Speech Bias Study. Values are in \%.}



\resizebox{0.6\columnwidth}{!}{%
\begin{tabular}{@{\extracolsep{-4pt}}lcccc}
\toprule
\makecell{\bfseries DN} &
\makecell{\bfseries \boldsymbol{$FPR_{1}$}} &
\makecell{\bfseries \boldsymbol{$FPR_{2}$}} & 
\makecell{\bfseries \boldsymbol{$FPR_{3}$}}& 
\makecell{\bfseries \boldsymbol{$EER$}} \\ 
\midrule
D01 & 88.10 & 96.54 & 66.13 & 34.10 \\
D02 & 69.96 & 96.77 & 24.99 & 18.16 \\

\midrule
D03 & 96.40 & 95.58 & 97.75 & 47.19 \\
D04  & 52.02 & 53.09 & 50.68 & 22.40 \\

\midrule
D05 & 97.32 & 98.53 & 94.06 & 49.05  \\
D06 & 87.44 & 83.41 & 92.18 & 46.94  \\
\midrule
Mean & 81.87 &87.32 &	70.97 &	36.31 \\
\bottomrule
\end{tabular}}
    \label{tab:stuttering_bias}
\end{table}
In this experiment, we examine the performance of existing synthetic speech detectors on bona fide speech from impaired speakers, particularly with stuttering.

From ~\cref{tab:results-asv2019}, the mean $EER$ of all the detectors on $D_{eval}$ set which contains bona fide speech from fluent speakers and synthetic speech from 11 unknown and 2 known speech generators is $5.29\%$.
However, results from~\cref{tab:stuttering_bias} show that the mean $EER$ from each detector increases and become $36.31$\% when the bona fide speech in $D_{eval}$ set is replaced with bona fide speech with stuttering from Sep-28k~\cite{sep_1, sep_2} dataset.
Note: The $EER$ rate is same as $FPR$ and $FNR$ as $EER$ is the rate that makes $FPR$ and $FNR$ equal.
Therefore, we can conclude that in the scenario where ground labels are given and $EER$ is calculated, the mean $FPR@EER \ \text{Threshold}$  would also increase from $5.29\%$ to $36.31\%$ for bona fide speech from impaired speakers (also shown in~\cref{fig:fpr}).

In the scenarios where ground truth labels are not provided, and $FPR$ is measured using pre-determined thresholds obtained from an independent dataset,
we can observe that mean $FPR$s are always greater than 70\%.
Therefore all detectors are biased and often misclassify bona fide speech with stuttering as synthetic more than fluent bona fide speech.

\section{Conclusions }\label{sec:conclusions}

In this work, we examined bias in different methods for synthetic speech detection. 
We processed a corpus of 1.7 million bona fide speech 
to create 28 different demographic sets.
We evaluated age, gender and accent bias in 6 synthetic speech detectors. 
Results indicate that 
synthetic speech detectors are 
biased.
False positive rates are higher for bona fide speech from male gender, speakers in age groups teens and 60s, and speakers with South Asian and Australian English accents in comparison to other demographic groups.
We also found that synthetic speech detectors are unfair to speech-impaired speakers.
Future work will focus on 
developing unbiased synthetic speech detectors.

\footnotesize
\noindent \textbf{Acknowledgements} This material is based on research sponsored by the Defense Advanced Research Projects Agency (DARPA) and the Air Force Research Laboratory (AFRL) under agreement number FA8750-20-2-1004. 
The U.S. Government is authorized to reproduce and distribute reprints for Governmental purposes notwithstanding any copyright notation thereon. The views and conclusions contained herein are those of the authors and should not be interpreted as necessarily representing the official policies or endorsements, either expressed or implied, of DARPA, AFRL or the U.S. Government. Address all correspondence to Edward J. Delp, \texttt{ace@purdue.edu}.
{\small
\bibliographystyle{IEEEtran}
\bibliography{ref}

\begin{thebibliography}{10}
\providecommand{\url}[1]{#1}
\csname url@samestyle\endcsname
\providecommand{\newblock}{\relax}
\providecommand{\bibinfo}[2]{#2}
\providecommand{\BIBentrySTDinterwordspacing}{\spaceskip=0pt\relax}
\providecommand{\BIBentryALTinterwordstretchfactor}{4}
\providecommand{\BIBentryALTinterwordspacing}{\spaceskip=\fontdimen2\font plus
\BIBentryALTinterwordstretchfactor\fontdimen3\font minus \fontdimen4\font\relax}
\providecommand{\BIBforeignlanguage}[2]{{%
\expandafter\ifx\csname l@#1\endcsname\relax
\typeout{** WARNING: IEEEtran.bst: No hyphenation pattern has been}%
\typeout{** loaded for the language `#1'. Using the pattern for}%
\typeout{** the default language instead.}%
\else
\language=\csname l@#1\endcsname
\fi
#2}}
\providecommand{\BIBdecl}{\relax}
\BIBdecl

\bibitem{asvspoof19}
M.~Todisco, X.~Wang, V.~Vestman, M.~Sahidullah, H.~Delgado, A.~Nautsch, J.~Yamagishi, N.~Evans, T.~Kinnunen, and K.~A. Lee, ``{ASVspoof 2019: Future Horizons in Spoofed and Fake Audio Detection},'' \emph{Proceedings of the ISCA Interspeech}, pp. 1008--1012, September 2019, {Graz, Austria}.

\bibitem{bhagtani2022overview}
K.~Bhagtani, A.~K.~S. Yadav, E.~R. Bartusiak, Z.~Xiang, R.~Shao, S.~Baireddy, and E.~J. Delp, ``{An Overview of Recent Work in Multimedia Forensics},'' \emph{Proceedings of the IEEE Conference on Multimedia Information Processing and Retrieval}, pp. 324--329, August 2022, {Virtual}.

\bibitem{ho2020}
J.~Ho, A.~Jain, and P.~Abbeel, ``{Denoising Diffusion Probabilistic Models},'' \emph{Advances in Neural Information Processing Systems}, vol.~33, pp. 6840--6851, December 2020.

\bibitem{dhariwal2021diffusion}
P.~Dhariwal and A.~Q. Nichol, ``{Diffusion Models Beat {GAN}s on Image Synthesis},'' \emph{Advances in Neural Information Processing Systems}, vol.~34, pp. 8780--8794, December 2021, {Virtual}.

\bibitem{stable_diffusion_cvpr2022}
R.~Rombach, A.~Blattmann, D.~Lorenz, P.~Esser, and B.~Ommer, ``{High-resolution image synthesis with latent diffusion models},'' \emph{Proceedings of the IEEE/CVF Conference on Computer Vision and Pattern Recognition}, pp. 10\,684--10\,695, 2022.

\bibitem{elevenlabs2023}
\BIBentryALTinterwordspacing
``{Speech Synthesis, ElevenLabs},'' December 2023. [Online]. Available: \url{https://elevenlabs.io/}
\BIBentrySTDinterwordspacing

\bibitem{unitspeech_interspeech2023}
H.~Kim, S.~Kim, J.~Yeom, and S.~Yoon, ``{UnitSpeech: Speaker-adaptive Speech Synthesis with Untranscribed Data},'' \emph{Proceedings of the ISCA Interspeech}, pp. 3038--3042, August 2023, {Dublin, Ireland}.

\bibitem{xttsv2}
\BIBentryALTinterwordspacing
Coqui, ``{XTTS},'' September 2023. [Online]. Available: \url{https://docs.coqui.ai/en/latest/models/xtts.html}
\BIBentrySTDinterwordspacing

\bibitem{huang2022prodiff}
R.~Huang, Z.~Zhao, H.~Liu, J.~Liu, C.~Cui, and Y.~Ren, ``{ProDiff: Progressive Fast Diffusion Model for High-Quality Text-to-Speech},'' \emph{Proceedings of the ACM International Conference on Multimedia}, pp. 2595--2605, October 2022, {Lisbon, Portugal}.

\bibitem{nyt_2023}
\BIBentryALTinterwordspacing
E.~Flitter and S.~Cowley, ``{Voice Deepfakes Are Coming for Your Bank Balance.}'' \emph{The New York Times}, August 2023. [Online]. Available: \url{https://www.nytimes.com/2023/08/30/business/voice-deepfakes-bank-scams.html}
\BIBentrySTDinterwordspacing

\bibitem{bi_2023}
\BIBentryALTinterwordspacing
B.~Nguyen, ``{A couple in Canada were reportedly scammed out of \$21,000 after getting a call from an AI-generated voice pretending to be their son.}'' \emph{The New York Times}, March 2023. [Online]. Available: \url{https://www.businessinsider.com/couple-canada-reportedly-lost-21000-in-ai-generated-voice-scam-2023-3}
\BIBentrySTDinterwordspacing

\bibitem{fortune_2023}
\BIBentryALTinterwordspacing
B.~MOLLMAN, ``{Scammers are using voice-cloning A.I. tools to sound like victims’ relatives in desperate need of financial help. It’s working.}'' \emph{The New York Times}, March 2023. [Online]. Available: \url{https://fortune.com/2023/03/05/scammers-ai-voice-cloning-tricking-victims-sound-like-relatives-needing-money/}
\BIBentrySTDinterwordspacing

\bibitem{wp_2023}
\BIBentryALTinterwordspacing
P.~Verma, ``{They thought loved ones were calling for help. It was an AI scam.}'' \emph{The Washington Post}, March 2023. [Online]. Available: \url{https://www.washingtonpost.com/technology/2023/03/05/ai-voice-scam/}
\BIBentrySTDinterwordspacing

\bibitem{tssdnet_2021}
G.~Hua, A.~B.~J. Teoh, and H.~Zhang, ``{Towards End-to-End Synthetic Speech Detection},'' \emph{IEEE Signal Processing Letters}, vol.~28, pp. 1265--1269, June 2021.

\bibitem{asvspoof_2021}
X.~Liu, X.~Wang, M.~Sahidullah, J.~Patino, H.~Delgado, T.~Kinnunen, M.~Todisco, J.~Yamagishi, N.~Evans, A.~Nautsch \emph{et~al.}, ``Asvspoof 2021: Towards spoofed and deepfake speech detection in the wild,'' \emph{arXiv preprint}, 2022.

\bibitem{tak22_odyssey}
H.~Tak, M.~Todisco, X.~Wang, J.~weon Jung, J.~Yamagishi, and N.~Evans, ``{Automatic Speaker Verification Spoofing and Deepfake Detection Using Wav2vec 2.0 and Data Augmentation},'' \emph{Proceedings of the Speaker and Language Recognition Workshop, Odyssey}, pp. 112--119, July 2022, {Beijing, China}.

\bibitem{add2023_submission_li2022convolutional}
K.~Li, X.-M. Zeng, J.-T. Zhang, and Y.~Song, ``Convolutional recurrent neural network and multitask learning for manipulation region location,'' \emph{Proceedings of IJCAI Workshop on Deepfake Audio Detection and Analysis}, pp. 18--22, August 2023, {Macao}.

\bibitem{acm_21_logspec}
Z.~Zhang, X.~Yi, and X.~Zhao, ``{Fake Speech Detection Using Residual Network with Transformer Encoder},'' \emph{Proceedings of the ACM Workshop on Information Hiding and Multimedia Security}, p. 13–22, June 2021, virtual Event, Belgium.

\bibitem{Sun_2023_CVPR}
C.~Sun, S.~Jia, S.~Hou, and S.~Lyu, ``Ai-synthesized voice detection using neural vocoder artifacts,'' \emph{Proceedings of the IEEE/CVF Conference on Computer Vision and Pattern Recognition Workshops}, pp. 904--912, June 2023, {Vancouver, Canada}.

\bibitem{alan_acm_2023}
Z.~Xiang, A.~K.~S. Yadav, S.~Tubaro, P.~Bestagini, and E.~J. Delp, ``Extracting efficient spectrograms from mp3 compressed speech signals for synthetic speech detection,'' \emph{Proceedings of the ACM Workshop on Information Hiding and Multimedia Security}, p. 163–168, 2023, {Chicago, IL, USA}.

\bibitem{asvdata_2019}
\BIBentryALTinterwordspacing
J.~Yamagishi, M.~Todisco, M.~Sahidullah, H.~Delgado, X.~Wang, N.~Evans, T.~Kinnunen, K.~Lee, V.~Vestman, and A.~Nautsch, ``{ASVspoof 2019: The 3rd Automatic Speaker Verification Spoofing and Countermeasures Challenge database},'' \emph{University of Edinburgh. The Centre for Speech Technology Research}, March 2019. [Online]. Available: \url{https://www.asvspoof.org/index2019.html}
\BIBentrySTDinterwordspacing

\bibitem{in_the_wild}
N.~M. M{\"u}ller, P.~Czempin, F.~Dieckmann, A.~Froghyar, and K.~B{\"o}ttinger, ``Does audio deepfake detection generalize?'' \emph{Proceedings of the ISCA Interspeech}, September 2022, {Incheon, Korea}.

\bibitem{ps3dt}
A.~K. Singh~Yadav, Z.~Xiang, K.~Bhagtani, P.~Bestagini, S.~Tubaro, and E.~J. Delp, ``{PS3DT: Synthetic Speech Detection Using Patched Spectrogram Transformer},'' \emph{Proceedings of the IEEE International Conference on Machine Learning and Applications}, pp. 496--503, 2023, {Florida, USA}.

\bibitem{dsvae_arXiv}
A.~K.~S. Yadav, K.~Bhagtani, Z.~Xiang, P.~Bestagini, S.~Tubaro, and E.~J. Delp, ``{DSVAE: Interpretable Disentangled Representation for Synthetic Speech Detection},'' \emph{{arXiv:2304.03323}}, April 2023.

\bibitem{exp_fake_22}
S.-Y. Lim, D.-K. Chae, and S.-C. Lee, ``{Detecting Deepfake Voice Using Explainable Deep Learning Techniques},'' \emph{Applied Sciences}, vol.~12, no.~8, 2022.

\bibitem{exp_cqcs_slt20}
H.~Tak, J.~Patino, A.~Nautsch, N.~Evans, and M.~Todisco, ``{An explainability study of the constant Q cepstral coefficient spoofing countermeasure for automatic speaker verification },'' \emph{Proceedings of the Speaker and Language Recognition Workshop}, pp. 333--340, November 2020, {Tokyo, Japan}.

\bibitem{salvi2023towards}
D.~Salvi, P.~Bestagini, and S.~Tubaro, ``Towards frequency band explainability in synthetic speech detection,'' in \emph{European Signal Processing Conference (EUSIPCO)}.\hskip 1em plus 0.5em minus 0.4em\relax IEEE, 2023.

\bibitem{yadav2023dsvae}
A.~K. Singh~Yadav, K.~Bhagtani, Z.~Xiang, P.~Bestagini, S.~Tubaro, and E.~J. Delp, ``{DSVAE: Disentangled Representation Learning for Synthetic Speech Detection},'' \emph{Proceedings of the IEEE International Conference on Machine Learning and Applications}, pp. 472--479, 2023, {Florida, USA}.

\bibitem{assd_2023}
A.~K. Singh~Yadav, Z.~Xiang, E.~R. Bartusiak, P.~Bestagini, S.~Tubaro, and E.~J. Delp, ``{ASSD: Synthetic Speech Detection in the AAC Compressed Domain},'' \emph{Proceedings of the IEEE International Conference on Acoustics, Speech, and Signal Processing}, pp. 1--5, June 2023, {Rhodes Island, Greece}.

\bibitem{bias_asv_interspeech_2021}
S.~Feng, O.~Kudina, B.~M. Halpern, and O.~Scharenborg, ``Quantifying bias in automatic speech recognition,'' \emph{arXiv preprint arXiv:2103.15122}, March 2021.

\bibitem{acm_bias_asv_2022}
W.~T. Hutiri and A.~Y. Ding, ``Bias in automated speaker recognition,'' \emph{Proceedings of the ACM Conference on Fairness, Accountability, and Transparency}, p. 230–247, June 2022, {Seoul, Republic of Korea}.

\bibitem{asr_inclusive_2022}
M.~K. Ngueajio and G.~Washington, ``{Hey ASR System! Why Aren't You More Inclusive?}'' \emph{Proceedings of International Conference on Human-Computer Interaction}, November 2022.

\bibitem{xu2023comprehensive}
Y.~Xu, P.~Terhörst, K.~Raja, and M.~Pedersen, ``A comprehensive analysis of ai biases in deepfake detection with massively annotated databases,'' \emph{arXiv preprint arXiv:2208.05845}, September 2023.

\bibitem{deepfake_fair_23}
M.~Pu, M.~Y. Kuan, N.~T. Lim, C.~Y. Chong, and M.~K. Lim, ``Fairness evaluation in deepfake detection models using metamorphic testing,'' \emph{Proceedings of the ACM International Workshop on Metamorphic Testing}, p. 7–14, January 2023, {Pittsburgh, Pennsylvania}.

\bibitem{mozilla_cvc}
\BIBentryALTinterwordspacing
M.~Organization, ``{Common Voice Corpus 16.1.}'' January 2024. [Online]. Available: \url{https://commonvoice.mozilla.org/en/datasets}
\BIBentrySTDinterwordspacing

\bibitem{sep_1}
C.~Lea, V.~Mitra, A.~Joshi, S.~Kajarekar, and J.~Bigham, ``Sep-28k: A dataset for stuttering event detection from podcasts with people who stutter,'' \emph{Proceedings of the IEEE International Conference on Acoustics, Speech and Signal Processing}, pp. 6798--6802, June 2021, {Toronto, Canada}.

\bibitem{sep_2}
S.~P. Bayerl, D.~Wagner, T.~Bocklet, and K.~Riedhammer, ``The {Influence} of {Dataset-Partitioning} on {Dysfluency} {Detection} {Systems},'' \emph{Text, {Speech}, and {Dialogue}}, vol. 13502, 2022.

\bibitem{mfccs}
M.~Sahidullah and G.~Saha, ``{Design, Analysis, and Experimental Evaluation of Block Based Transformation in MFCC Computation for Speaker Recognition},'' \emph{Speech Communication}, vol.~54, pp. 543--565, May 2012.

\bibitem{cqccs}
M.~Todisco, H.~Delgado, and N.~Evans, ``{Constant Q Cepstral Coefficients: A Spoofing Countermeasure for Automatic Speaker Verification},'' \emph{Computer Speech \& Language}, vol.~45, pp. 516--535, September 2017.

\bibitem{Borzi_2022_CVPR}
S.~Borz{\`\i}, O.~Giudice, F.~Stanco, and D.~Allegra, ``Is synthetic voice detection research going into the right direction?'' \emph{Proceedings of the IEEE/CVF Conference on Computer Vision and Pattern Recognition Workshops}, pp. 71--80, June 2022, {New Orleans, USA}.

\bibitem{li2021replay}
X.~Li, N.~Li, C.~Weng, X.~Liu, D.~Su, D.~Yu, and H.~Meng, ``{Replay and Synthetic Speech Detection with Res2Net Architecture},'' \emph{Proceedings of the IEEE International Conference on Acoustics, Speech and Signal Processing}, pp. 6354--6358, June 2021, {Toronto, Canada}.

\bibitem{lfcc_interspeech}
M.~Sahidullah, T.~Kinnunen, and C.~Hanilçi, ``{A comparison of features for synthetic speech detection},'' \emph{Proceedings of the ISCA Interspeech}, pp. 2087--2091, September 2015, {Dresden, Germany}.

\bibitem{akdeniz2021detection}
F.~Akdeniz and Y.~Becerikli, ``{Detection of Copy-Move Forgery in Audio Signal with Mel Frequency and Delta-Mel Frequency Kepstrum Coefficients},'' \emph{Proceedings of the Innovations in Intelligent Systems and Applications Conference}, pp. 1--6, October 2021, {Elazig, Turkey}.

\bibitem{alzantot2019}
M.~Alzantot, Z.~Wang, and M.~B. Srivastava, ``{Deep Residual Neural Networks for Audio Spoofing Detection},'' \emph{Proceedings of the ISCA Interspeech}, pp. 1078--1082, September 2019, {Graz, Austria}.

\bibitem{he2016deep}
K.~He, X.~Zhang, S.~Ren, and J.~Sun, ``{Deep Residual Learning for Image Recognition},'' \emph{Proceedings of the IEEE Conference on Computer Vision and Pattern Recognition}, pp. 770--778, June 2016, {Las Vegas, NV}.

\bibitem{spec_vgg_sincnet_28}
H.~Zeinali, T.~Stafylakis, G.~Athanasopoulou, J.~Rohdin, I.~Gkinis, L.~Burget, and J.~{\v{C}}ernock{\`y}, ``{Detecting Spoofing Attacks Using VGG and SincNet: BUT-Omilia Submission to ASVspoof 2019 Challenge},'' \emph{Proceedings of the ISCA Interspeech}, pp. 1073--1077, {September} 2019, {Graz, Austria}.

\bibitem{bartusiak2023}
E.~R. Bartusiak, K.~Bhagtani, A.~K.~S. Yadav, and E.~J. Delp, ``{Transformer Ensemble for Synthesized Speech Detection},'' \emph{Proceedings of the Asilomar Conference on Signals, Systems, and Computers}, October 2023, {Pacific Grove, California, USA}.

\bibitem{mel}
S.~S. Stevens, J.~Volkmann, and E.~B. Newman, ``{A Scale for the Measurement of the Psychological Magnitude Pitch},'' \emph{Journal of the Acoustical Society of America}, vol.~8, pp. 185--190, June 1937.

\bibitem{vae_first_paper_2013}
D.~P. Kingma and M.~Welling, ``Auto-encoding variational bayes,'' \emph{arXiv preprint arXiv:1312.6114}, 2013.

\bibitem{vaswani_2017}
A.~Vaswani, N.~Shazeer, N.~Parmar, J.~Uszkoreit, L.~Jones, A.~N. Gomez, Łukasz Kaiser, and I.~Polosukhin, ``{Attention is All You Need},'' \emph{Proceedings of the Neural Information Processing Systems}, December 2017, {Long Beach, CA}.

\bibitem{koutini_2021}
K.~Koutini, J.~Schlüter, H.~Eghbal-zadeh, and G.~Widmer, ``{Efficient Training of Audio Transformers with Patchout},'' \emph{Proceedings of the ISCA Interspeech}, pp. 2753--2757, September 2022, {Incheon, Korea}.

\bibitem{gong_2021_ssast}
Y.~Gong, C.-I. Lai, Y.-A. Chung, and J.~Glass, ``{SSAST: Self-Supervised Audio Spectrogram Transformer},'' \emph{Proceedings of the AAAI Conference on Artificial Intelligence}, vol.~36, no.~10, pp. 10\,699--10\,709, October 2022, {Virtual}.

\bibitem{gong_2021_ast}
Y.~Gong, Y.-A. Chung, and J.~Glass, ``{AST: Audio Spectrogram Transformer},'' \emph{Proceedings of the ISCA Interspeech}, pp. 571--575, August 2021, {Brno, Czech Republic}.

\bibitem{amit_ei_paper}
A.~K.~S. Yadav, E.~Bartusiak, K.~Bhagtani, and E.~J. Delp, ``{Synthetic Speech Attribution using Self Supervised Audio Spectrogram Transformer},'' \emph{Proceedings of the IS\&T Media Watermarking, Security, and Forensics Conference, Electronic Imaging Symposium}, January 2023, san Francisco, CA.

\bibitem{acm_kratika_2023}
K.~Bhagtani, E.~R. Bartusiak, A.~K.~S. Yadav, P.~Bestagini, and E.~J. Delp, ``Synthesized speech attribution using the patchout spectrogram attribution transformer,'' \emph{Proceedings of the ACM Workshop on Information Hiding and Multimedia Security}, p. 157–162, June 2023, {Chicago, IL, USA}.

\bibitem{mdrt_amit}
A.~K. Singh~Yadav, K.~Bhagtani, S.~Baireddy, P.~Bestagini, S.~Tubaro, and E.~J. Delp, ``Mdrt: Multi-domain synthetic speech localization,'' \emph{Proceedings of IEEE International Conference on Acoustics, Speech and Signal Processing}, pp. 11\,171--11\,175, 2024, {Seoul, South Korea}.

\bibitem{fgsat_2023}
K.~Bhagtani, A.~K.~S. Yadav, Z.~Xiang, P.~Bestagini, and E.~J. Delp, ``{FGSSAT : Unsupervised Fine-Grain Attribution of Unknown Speech Synthesizers Using Transformer Networks},'' \emph{Proceedings of the IEEE Asilomar Conference on Signals, Systems, and Computers}, pp. 1135--1140, 2023, {Pacific Grove, CA}.

\bibitem{multi_task_zhang21_asvspoof}
L.~Zhang, X.~Wang, E.~Cooper, and J.~Yamagishi, ``{Multi-task Learning in Utterance-level and Segmental-level Spoof Detection},'' \emph{Proceedings of the Edition of the Automatic Speaker Verification and Spoofing Countermeasures Challenge}, pp. 9--15, September 2021, {Online}.

\bibitem{acm_local_attention_paper_2023}
Y.~Zhu, Y.~Chen, Z.~Zhao, X.~Liu, and J.~Guo, ``Local self-attention based hybrid multiple instance learning for partial spoof speech detection,'' \emph{ACM Transactions on Intelligent Systems and Technology}, August 2023.

\bibitem{wav2vec2_2020}
A.~Baevski, H.~Zhou, A.~Mohamed, and M.~Auli, ``wav2vec 2.0: A framework for self-supervised learning of speech representations,'' October 2020.

\bibitem{PartialSpoof}
L.~Zhang, X.~Wang, E.~Cooper, N.~Evans, and J.~Yamagishi, ``The partialspoof database and countermeasures for the detection of short fake speech segments embedded in an utterance,'' \emph{IEEE/ACM Transactions on Audio, Speech, and Language Processing}, vol.~31, pp. 813--825, 2023.

\bibitem{audioset}
J.~F. Gemmeke, D.~P. Ellis, D.~Freedman, A.~Jansen, W.~Lawrence, R.~C. Moore, M.~Plakal, and M.~Ritter, ``{Audio set: An Ontology and Human-labeled Dataset for Audio Events},'' \emph{Proceedings of the IEEE International Conference on Acoustics, Speech and Signal Processing}, March 2017, {New Orleans, LA}.

\bibitem{librispeech}
V.~Panayotov, G.~Chen, D.~Povey, and S.~Khudanpur, ``{LibriSpeech: an ASR Corpus Based on Public Domain Audio Books},'' \emph{Proceedings of the IEEE International Conference on Acoustics, Speech and Signal Processing}, April 2015, {Queensland, Australia}.

\bibitem{exp_fake_det_icassp22}
W.~Ge, J.~Patino, M.~Todisco, and N.~Evans, ``{Explaining Deep Learning Models for Spoofing and Deepfake Detection with Shapley Additive Explanations},'' \emph{Proceedings of the IEEE International Conference on Acoustics, Speech and Signal Processing}, pp. 6387--6391, May 2022, {Singapore}.

\bibitem{wacv_21_fairface}
K.~Kärkkäinen and J.~Joo, ``Fairface: Face attribute dataset for balanced race, gender, and age for bias measurement and mitigation,'' \emph{Proceedings of IEEE Winter Conference on Applications of Computer Vision}, pp. 1547--1557, January 2021, {Hawaii, USA}.

\bibitem{21_survey_bias_fairness}
N.~Mehrabi, F.~Morstatter, N.~Saxena, K.~Lerman, and A.~Galstyan, ``A survey on bias and fairness in machine learning,'' \emph{ACM Computing Surveys}, vol.~54, no.~6, pp. 1--35, July 2021.

\bibitem{22_tbiom_ai_fair}
C.~Hazirbas, J.~Bitton, B.~Dolhansky, J.~Pan, A.~Gordo, and C.~C. Ferrer, ``Towards measuring fairness in ai: The casual conversations dataset,'' \emph{IEEE Transactions on Biometrics, Behavior, and Identity Science}, vol.~4, no.~3, pp. 324--332, 2022.

\bibitem{open_challenge_23}
M.~Masood, M.~Nawaz, K.~M. Malik, A.~Javed, A.~Irtaza, and H.~Malik, ``Deepfakes generation and detection: state-of-the-art, open challenges, countermeasures, and way forward,'' \emph{Applied Intelligence}, vol.~53, no.~4, pp. 3974--4026, June 2023.

\bibitem{wacv_w_22}
R.~Ramachandra, K.~Raja, and C.~Busch, ``Algorithmic fairness in face morphing attack detection,'' \emph{Proceedings of IEEE/CVF Winter Conference on Applications of Computer Vision Workshops}, pp. 410--418, January 2022, {Hawaii, USA}.

\bibitem{fairness_face_2023}
M.~Fang, W.~Yang, A.~Kuijper, V.~Struc, and N.~Damer, ``Fairness in face presentation attack detection,'' \emph{Pattern Recognition}, vol. 147, p. 110002, October 2023.

\bibitem{trinh2021examination}
L.~Trinh and Y.~Liu, ``An examination of fairness of ai models for deepfake detection,'' \emph{Proceedings of the International Joint Conference on Artificial Intelligence}, pp. 567--574, August 2021, {Montreal, Canada}.

\bibitem{nadimpalli2022gbdf}
A.~V. Nadimpalli and A.~Rattani, ``Gbdf: Gender balanced deepfake dataset towards fair deepfake detection,'' \emph{arXiv preprint arXiv:2207.10246}, July 2022.

\bibitem{Ju_2024_WACV}
Y.~Ju, S.~Hu, S.~Jia, G.~H. Chen, and S.~Lyu, ``Improving fairness in deepfake detection,'' \emph{Proceedings of the IEEE Winter Conference on Applications of Computer Vision}, pp. 4655--4665, January 2024, {Hawaii, USA}.

\bibitem{msm_mae_niizumi}
D.~Niizumi, D.~Takeuchi, Y.~Ohishi, N.~Harada, and K.~Kashino, ``{Masked Spectrogram Modeling using Masked Autoencoders for Learning General-purpose Audio Representation},'' \emph{Proceedings of Machine Learning Research}, vol. 166, pp. 1--24, Dec 2022.

\bibitem{aasist}
J.-w. Jung, H.-S. Heo, H.~Tak, H.-j. Shim, J.~S. Chung, B.-J. Lee, H.-J. Yu, and N.~Evans, ``Aasist: Audio anti-spoofing using integrated spectro-temporal graph attention networks,'' \emph{Proceedings of the IEEE International Conference on Acoustics, Speech and Signal Processing}, pp. 6367--6371, May 2022, {Singapore}.

\bibitem{rawpc_darts}
W.~Ge, J.~Patino, M.~Todisco, and N.~Evans, ``{Raw Differentiable Architecture Search for Speech Deepfake and Spoofing Detection},'' \emph{Proceedings of the ISCA Automatic Speaker Verification and Spoofing Countermeasures Challenge}, pp. 22--28, 2021.

\bibitem{asvspoof19_baselines}
A.~Nautsch, X.~Wang, N.~Evans, T.~H. Kinnunen, V.~Vestman, M.~Todisco, H.~Delgado, M.~Sahidullah, J.~Yamagishi, and K.~A. Lee, ``{ASVspoof 2019: Spoofing Countermeasures for the Detection of Synthesized, Converted and Replayed Speech},'' \emph{IEEE Transactions on Biometrics, Behavior, and Identity Science}, vol.~3, no.~2, pp. 252--265, February 2021.

\bibitem{ps3dt_arXiv}
A.~K.~S. Yadav, Z.~Xiang, K.~Bhagtani, P.~Bestagini, S.~Tubaro, and E.~J. Delp, ``{Compression Robust Synthetic Speech Detection Using Patched Spectrogram Transformer},'' \emph{arXiv:2402.14205}, February 2024.

\bibitem{loshchilov_2019}
I.~Loshchilov and F.~Hutter, ``{Decoupled Weight Decay Regularization},'' \emph{Proceedings of the International Conference on Learning Representations}, May 2019, {New Orleans, LA}.

\bibitem{Tange2011a}
\BIBentryALTinterwordspacing
O.~Tange, ``Gnu parallel - the command-line power tool,'' \emph{;login: The USENIX Magazine}, vol.~36, no.~1, pp. 42--47, Feb 2011. [Online]. Available: \url{http://www.gnu.org/s/parallel}
\BIBentrySTDinterwordspacing

\end{thebibliography}
}

\clearpage
\newpage
\onecolumn

\appendix

\setcounter{table}{0}
\renewcommand{\thetable}{A\arabic{table}}
\setcounter{figure}{0}
\renewcommand{\thefigure}{A\arabic{figure}}

\twocolumn[{\centering{\huge Supplementary Material \\ 
\huge FairSSD: Understanding Bias in Synthetic Speech Detectors
\par}\vspace{3ex}}
\smallbreak]

\section{Detectors Training Details}\label{sec:sup_train-details}
Below we provide additional training details for the detectors used in our study.
\subsection{LFCC-GMMs}
While this method in~\cite{asvspoof19} is implemented in MATLAB, we used scikit learn to fit the~\gls{gmms}. 
~\cref{tab:sup_gmms} presents our ablation study.
The M01 version reports the performance of the method trained in MATLAB using the parameters provided for the baseline in ASVspoof2019 Challenge~\cite{asvspoof19_baselines}.
M01 processes 20ms windows with a hop size of 10ms and frequency components from 30Hz to 8kHz to obtain \gls{lfccs} features.
The M02 version reports the performance of the method using same parameters but trained in Python using scikit learn.
The M03 version is implemented in scikit learn and processes 30ms windows with a hop size of 15ms and frequency components upto 4kHz to obtain \gls{lfccs} features.
\cref{tab:sup_gmms} shows the~\gls{eer} in \% of all three versions on the $D_{eval}$ set of the ASVspoof2019 dataset.
Two~\gls{gmms} are trained separately, each one for the bona fide and synthetic classes using Expectation Maximization (EM) algorithm.
Each GMM has 512 Gaussian mixture components.
The covariance matrix of the Gaussian distributions was constrained to be a diagonal matrix.
For all three versions, we obtained 20~\gls{lfccs}, their deltas and also delta-delta coefficients as described in~\cite{asvspoof19_baselines}.
Therefore, we obtain 60-dimensional features for each small window of the speech signal (20ms in case of M01 and M02, and 30ms in case of M03).
Since window size in M02 is smaller, we obtained a higher number of windows in case of M02 as compared to that of M03.
We observed that M02 was relatively more computationally intensive than M03, and was slow to train but both versions had comparable performance on the evaluation set of the ASVspoof2019 dataset as shown in ~\cref{tab:sup_gmms}.
In our study, we required to evaluate this method on 0.9 million speech signals, which is why we selected M03 for our study to examine bias in LFCC-GMMs~\cite{asvspoof19, lfcc_interspeech} synthetic speech detector.

\begin{table}[!h]
    \centering
    \caption{Ablation on LFCC-GMMs showing~\gls{eer} in \% on $D_{eval}$.}
    \resizebox{\columnwidth}{!}{
\begin{tabular}{@{\extracolsep{-4pt}}lcccc}
\toprule
\makecell{ \bfseries Version} & 
\makecell{\bfseries Window Size}&
\makecell{\bfseries Hop Size} &
\makecell{\bfseries Package} &  
\makecell{\bfseries \Vec{$D_{eval}$}} \\ 
\midrule
M01 & {20ms} & 10ms & MATLAB & 8.09\\
M02 & {20ms}  & 10ms & scikit-learn & 3.67 \\
M03 & {30ms} & 15ms & scikit-learn & 3.46\\
\bottomrule
\end{tabular}}
    \label{tab:sup_gmms}
\end{table}

\subsection{MFCC-ResNet}

\begin{table}[!h]
    \centering
    \caption{Ablation on MFCC-ResNet showing~\gls{eer} in \% on $D_{dev}$ and $D_{eval}$.}
    \resizebox{\columnwidth}{!}{%
\begin{tabular}{@{\extracolsep{-4pt}}lcc}
\toprule
\makecell{\bfseries Method \bfseries Version} &  
\makecell{\bfseries \Vec{$D_{dev}$}} & 
\makecell{\bfseries \Vec{$D_{eval}$}} \\ 
\midrule
Presented in~\cite{alzantot2019} & 3.34 & 9.33\\
Using provided model weights & >40.00 & >40.00\\
Our retrained version & 6.52 & 11.58 \\
\bottomrule
\end{tabular}}
    \label{tab:sup_mfcc-resnet}
\end{table}

The features used by this detector consist of~\gls{mfcc}, and its first and second derivatives. 
The~\gls{mfcc} are obtained using the~\gls{stft} of the speech signal, mel-spectrum filters and ~\gls{dct}~\cite{alzantot2019}.
We select the first 24 coefficients.
Including, its first and second derivatives, in total, it leads to a feature of dimension 72. 
We use these parameters because they have the best performance in ~\cite{alzantot2019}.
These hand-crafted features are processed by a ResNet~\cite{he2016deep} provided in~\cite{alzantot2019} for synthetic speech detection.

The weights provided with~\cite{alzantot2019} were for a model which was trained on~\gls{mfcc} features obtained using the librosa.feature package.
However, later (after the source code and model weights were released), the package was updated
which changed the number of windows obtained for each speech signal. 
Hence, corresponding changes were needed to be made in the ResNet architecture provided by the authors in the source code.
After making these number of windows changes, in our first experiment, we evaluated the performance of the model weights provided by the authors.
There was a significant drop in performance as compared to the results presented in~\cite{alzantot2019}. 
Using the model weights provided by the authors, we observed an \gls{eer} higher than 40\% on both development $D_{dev}$ and evaluation $D_{eval}$ sets of the ASVspoof2019 Dataset.
Hence, we retrained the method on the ASVspoof2019 dataset and obtained our own model weights.

We trained this method for 200 epochs using a learning rate of \(5 \times 10^{-5}\) and a batch size of 32.
We selected the model which performed the best on the validation set of ASVspoof2019.
In ~\cref{tab:sup_mfcc-resnet}, the \gls{eer} in \% on $D_{dev}$ and $D_{eval}$ sets of the ASVspoof2019 dataset obtained by using our own model weights are shown.
This led to better performance than directly using the model weights provided by the authors, hence we used the re-trained version for our bias study.

\subsection{Spec-ResNet}
\begin{table}[!h]
    \centering
    \caption{Ablation on Spec-ResNet showing~\gls{eer} in \% on $D_{dev}$ and $D_{eval}$.}
    \resizebox{\columnwidth}{!}{%
\begin{tabular}{@{\extracolsep{-4pt}}lcc}
\toprule
\makecell{\bfseries Method \bfseries Version} &  
\makecell{\bfseries \Vec{$D_{dev}$}} & 
\makecell{\bfseries \Vec{$D_{eval}$}} \\ 
\midrule
Presented in~\cite{alzantot2019} & 0.11 & 9.68\\
Using provided model weights & 43.05 & 42.01\\
Our retrained version & 0.71 & 10.10 \\
\bottomrule
\end{tabular}}
    \label{tab:sup_spec-resnet}
\end{table}
For similar reasons as mentioned for MFCC-ResNet, we re-trained this method.
This method obtains a 2048-point~\gls{stft} from the speech signal using a window size of 2048 and hop length of 512.
Next, the squares of absolute value of the~\gls{stft} are obtained and converted into decibles (dB) scale, which is a logarithmic scale. 
We trained this method for 200 epochs using using a learning rate of \(5 \times 10^{-5}\) and a batch size of 32.
The model with the best performance on the validation set was selected.
The \gls{eer} in \% on $D_{dev}$ and $D_{eval}$ sets of ASVspoof2019 obtained using model weights provided in ~\cite{alzantot2019} and our re-trained model are shown in~\cref{tab:sup_spec-resnet}. 
We observed that our re-training helped to reduce \gls{eer} significantly. Hence, we use our retrained model weights for the bias study.
\begin{figure*}[!t]
    \centering
    \includegraphics[width=1.0\linewidth]{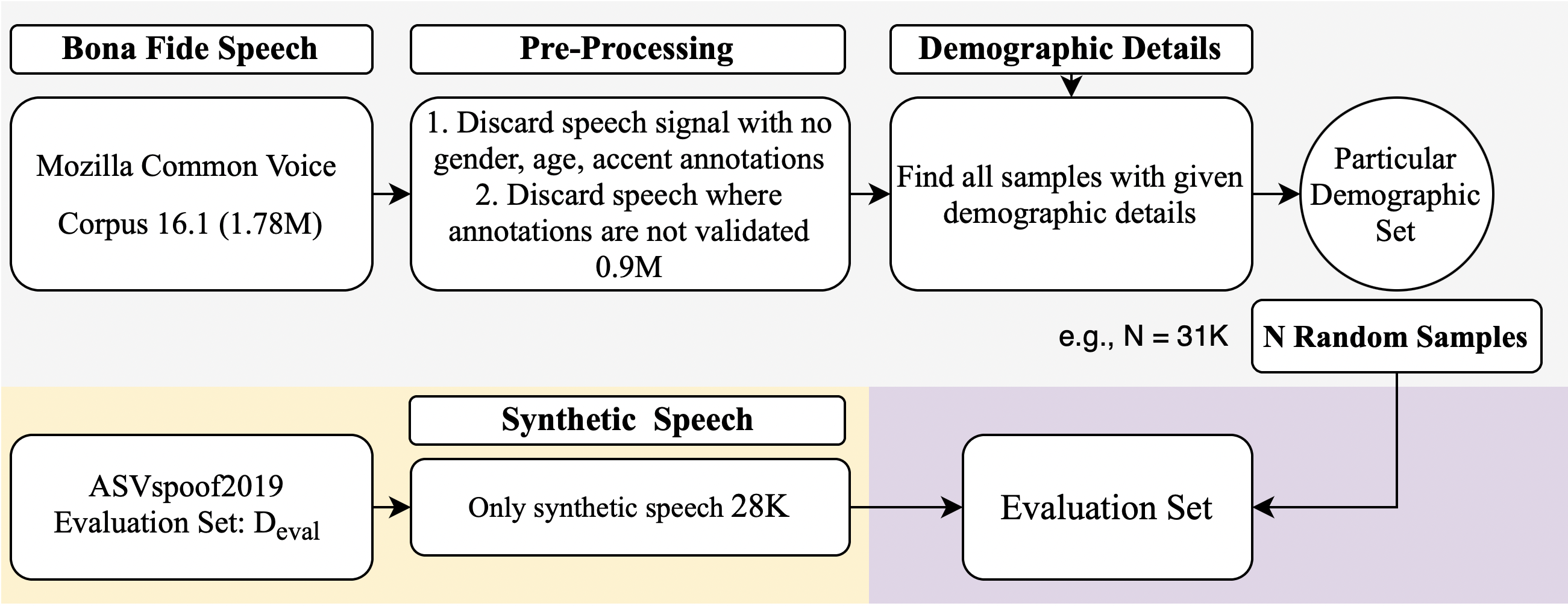}
    \caption{Overview of Dataset Preparation for bias study.}
    \label{fig:sup_dataset}
\end{figure*}
\subsection{PS3DT} 
To obtain mel-spectrogam, we used a Hanning window of size 25 ms with a shift of 10 ms. 
As mentioned in ~\cite{ps3dt_arXiv}, we used $80$ frequency bins and fixed input speech signal to $5.12$ seconds, resulting in a mel-spectrogram of size $80\times 512$.
The network was trained for 50 epochs with a batch size of 256 and AdamW optimizer~\cite{loshchilov_2019}.
The initial learning rate was set to $10^{-5}$ and a weight decay of $10^{-4}$ was used.
We selected the model weights which provide best accuracy on the validation set.
We obtained performance same as mentioned in ~\cite{ps3dt}.
\subsection{TSSDNet} 
We do not perform any re-training for this method. We performed evaluation using the ResNet style \gls{tssdnet} model weights provided in~\cite{tssdnet_2021} as the results obtained with it were same as reported in~\cite{tssdnet_2021}.
\subsection{Wav2Vec2-AASIST}
We do not perform any re-training for this method. We used the weights provided by the authors in~\cite{tak22_odyssey} for this method as the results obtained with it were same as reported in~\cite{tak22_odyssey}.

\section{Dataset Collection and Pre-Processing }\label{sec:sup_dataset}

In this section, we provide details about how we collected dataset for our bias study and how we processed it. 
For our age, gender and bias studies, we created 28 evaluation sets.
Each evaluation set has bona fide class and synthetic class as shown in~\cref{fig:sup_dataset}.
We kept synthetic class same in each set. 
The sets only differ in terms of bona fide speech signals and particularly bona fide speakers' demographics.
For collecting samples for bona fide class we downloaded the Mozilla Common Voice Corpus 16.1~\cite{mozilla_cvc}.
It has approximately 1.78 million English speech signals.
We pre-processed the dataset and obtained approximately 0.9 million speech signals having all the required annotations for our study.
To pre-process, we used the annotations provided in the dataset. 
To handle large number of files and its processing, we run parallel process using GNU parallel~\cite{Tange2011a}.
We used all the speech signal with gender, age and accent annotations.
We considered only annotations which are validated.
Also, since each detector is trained on 16KHz samples. 
We resampled each speech signal from the Mozilla Common Voice Corpus (mostly 44KHz) to 16KHz.

For gender, there are three categories: male, female and others in the Mozilla Common Voice Corpus.
In gender bias study, we fixed the accent to US English as it had most number of samples available. 
We studied three most frequent age groups in the dataset, namely, 20s, 30s and 60s.
For each group, we made two sets: one for male and other for female.
We kept same number of samples for each gender in gender bias study.
We limited our categories to only male and female as adding other gender samples lead to strong unbalance in the dataset and for fair evaluation, we wanted to keep number of samples same for both genders in a particular age group.

For age, we fixed accent to US English and the age categories in Mozilla Common Voice Corpus are teens, 20s, 30s, 40s, 50s, 60s, 70s, 80s and 90s. 
We studied gender bias in both male and female gender. 
For a fixed gender, we kept number of samples same for all age groups.
We discarded age groups 70s, 80s, and 90s as they had limited number of samples.

For accent, we fixed age group to 20s as it has the most number of samples. We examined accent bias for male and female gender separately.
There are more than 100 different labels for accents. However, many labels have even less than 100 speech samples. In our study, we selected 5 most frequent accents, namely, US English, Canadian English, British English, Australian English, South Asian English.
The number of samples are kept same for all the accents.
Note, in future work, we plan to provide more fine-grained analysis \eg within bias among accents with US English like Midwestern Accent, California Accent, and so on.

For each evaluation set, we first obtain a bigger set consisting of bona fide speech with a particular demographic as shown in~\cref{fig:sup_dataset}.
For example, in our gender study for demographic: speakers in 20s with US Accent, we get a bigger set with 31,500 speech signals for female and 109,000 speech signals for male. 
From both sets, we randomly sample `N' (in this case: `N' = 31K) speech signals as shown in ~\cref{fig:sup_dataset}.
This brings randomness in our experiment, therefore we report mean metrics and standard deviation obtained from 5 runs of each experiment.
For a given demographic, higher standard deviation will show that the results are highly dependent on the content than the demographic itself.
We obtained less than $0.05\%$ standard deviation for all metrics.
This indicates that the detectors and the results are not dependent on content but demographics in all our experiments.
The 28 evaluation datasets used in our age, gender and bias study and script we created to process the dataset can be found at \url{https://gitlab.com/viper-purdue/fairssd}. 
We believe this will help future research in this direction.

\section{Absolute Value of FPR and EER} \label{sec:sup_result}
In this section we provide the absolute values of all the metrics. 
Notice, each experiment is repeated 5 times to get the value. 
We report both mean and standard deviation (SD).
\cref{tab:sup_gender_bias} shows our gender bias study results.
Note: D01 is TSSDNet~\cite{tssdnet_2021}, D02 is Wav2Vec2-AASIST (Wav2Vec2)~\cite{tak22_odyssey}, D03 is Spec-ResNet~\cite{alzantot2019}, D04 is PS3DT~\cite{ps3dt}, D05 is LFCC-GMMs~\cite{asvspoof19_baselines, asvspoof19}, and D06 is MFCC-ResNet~\cite{alzantot2019}.
Similar to the results reported in the paper using difference, we can notice that $FPR$s are higher for males than for female counterparts for most detectors.
Similarly, in~\cref{tab:sup_age_m} and~\cref{tab:sup_age_f} we report our age bias study results for male and female genders.
Most detectors have higher bias ($FPR$ and $EER$) for people in age groups teens and 60s.
Finally, in~\cref{tab:sup_accent_m} and~\cref{tab:sup_accent_f}, we report results for accents. Most detectors have higher $FPR$s and $EER$s for speakers with South Asian and Australian English accents.

Notice, while the $\Delta FPR$s and $\Delta EER$ reported in the paper do not reveal actual performance, 
the absolute values reported here reveal absolute performance. 
We observe that LFFC-GMMs \ie, D05 has $100\%$ $FPR$s with threshold estimated from an independent dataset \ie $D_{eval}$ set of ASVspoof2019.
Hence, this method does not generalize on bona fide speech from unknown speakers and always misclassifies them as synthetic.
However, the $EER$ can still show bias. 
As both classes in each evaluation set have same results on synthetic class (as synthetic class is same in each set). Hence, higher $EER$ for one demographic group indicates method outputs higher probability for bona fide speech from one demographic than that for bona fide speech from other demographic group.

\begin{table*}
    \centering
    \caption{Absolute performance of detectors in gender bias study.}
    \resizebox{1.0\textwidth}{!}{

\begin{tabular}{@{}lllll|lllll|lllll@{}}
\textbf{Method}       & \textbf{Data}  & \textbf{Metric} & \textbf{Mean} & \textbf{SD} & \textbf{} & \textbf{Data}  & \textbf{Metric} & \textbf{Mean} & \textbf{SD} & \textbf{} & \textbf{Data}  & \textbf{Metric} & \textbf{Mean} & \textbf{SD} \\
\toprule
                      & $D_{US-20s-M}$ & $FPR_{1}$       & 98.26\%       & 0.086\%      &           & $D_{US-30s-M}$ & $FPR_{1}$       & 98.59\%       & 0.032\%      &           & $D_{US-60s-M}$ & $FPR_{1}$       & 98.53\%       & 0.013\%      \\
                      & $D_{US-20s-F}$ & $FPR_{1}$       & 96.79\%       & 0.008\%      &           & $D_{US-30s-F}$ & $FPR_{1}$       & 97.67\%       & 0.013\%      &           & $D_{US-60s-F}$ & $FPR_{1}$       & 91.97\%       & 0.075\%      \\
D01         & $D_{US-20s-M}$ & $FPR_{2}$       & 99.95\%       & 0.007\%      &           & $D_{US-30s-M}$ & $FPR_{2}$       & 99.98\%       & 0.011\%      &           & $D_{US-60s-M}$ & $FPR_{2}$       & 99.91\%       & 0.000\%      \\
                      & $D_{US-20s-F}$ & $FPR_{2}$       & 99.59\%       & 0.005\%      &           & $D_{US-30s-F}$ & $FPR_{2}$       & 99.88\%       & 0.004\%      &           & $D_{US-60s-F}$ & $FPR_{2}$       & 99.94\%       & 0.006\%      \\
                      & $D_{US-20s-M}$ & $FPR_{3}$       & 82.59\%       & 0.157\%      &           & $D_{US-30s-M}$ & $FPR_{3}$       & 83.26\%       & 0.273\%      &           & $D_{US-60s-M}$ & $FPR_{3}$       & 91.89\%       & 0.034\%      \\
                      & $D_{US-20s-F}$ & $FPR_{3}$       & 81.78\%       & 0.038\%      &           & $D_{US-30s-F}$ & $FPR_{3}$       & 77.25\%       & 0.071\%      &           & $D_{US-60s-F}$ & $FPR_{3}$       & 65.77\%       & 0.246\%      \\
                      & $D_{US-20s-M}$ & $EER$           & 46.57\%       & 0.050\%      &           & $D_{US-30s-M}$ & $EER$           & 43.25\%       & 0.170\%      &           & $D_{US-60s-M}$ & $EER$           & 57.88\%       & 0.020\%      \\
                      & $D_{US-20s-F}$ & $EER$           & 45.12\%       & 0.018\%      &           & $D_{US-30s-F}$ & $EER$           & 44.46\%       & 0.017\%      &           & $D_{US-60s-F}$ & $EER$           & 43.04\%       & 0.059\%      \\
                      &                &                 &               &              &           &                &                 &               &              &           &                &                 &               &              \\
                      & $D_{US-20s-M}$ & $FPR_{1}$       & 27.04\%       & 0.046\%      &           & $D_{US-30s-M}$ & $FPR_{1}$       & 21.36\%       & 0.310\%      &           & $D_{US-60s-M}$ & $FPR_{1}$       & 28.71\%       & 0.043\%      \\
                      & $D_{US-20s-F}$ & $FPR_{1}$       & 29.13\%       & 0.033\%      &           & $D_{US-30s-F}$ & $FPR_{1}$       & 23.76\%       & 0.036\%      &           & $D_{US-60s-F}$ & $FPR_{1}$       & 17.31\%       & 0.204\%      \\
D02 & $D_{US-20s-M}$ & $FPR_{2}$       & 90.70\%       & 0.131\%      &           & $D_{US-30s-M}$ & $FPR_{2}$       & 83.45\%       & 0.379\%      &           & $D_{US-60s-M}$ & $FPR_{2}$       & 93.10\%       & 0.032\%      \\
                      & $D_{US-20s-F}$ & $FPR_{2}$       & 91.31\%       & 0.018\%      &           & $D_{US-30s-F}$ & $FPR_{2}$       & 90.19\%       & 0.034\%      &           & $D_{US-60s-F}$ & $FPR_{2}$       & 92.72\%       & 0.096\%      \\
                      & $D_{US-20s-M}$ & $FPR_{3}$       & 2.67\%        & 0.112\%      &           & $D_{US-30s-M}$ & $FPR_{3}$       & 2.29\%        & 0.090\%      &           & $D_{US-60s-M}$ & $FPR_{3}$       & 1.47\%        & 0.016\%      \\
                      & $D_{US-20s-F}$ & $FPR_{3}$       & 2.35\%        & 0.014\%      &           & $D_{US-30s-F}$ & $FPR_{3}$       & 1.35\%        & 0.029\%      &           & $D_{US-60s-F}$ & $FPR_{3}$       & 0.69\%        & 0.009\%      \\
                      & $D_{US-20s-M}$ & $EER$           & 3.87\%        & 0.038\%      &           & $D_{US-30s-M}$ & $EER$           & 3.66\%        & 0.080\%      &           & $D_{US-60s-M}$ & $EER$           & 3.03\%        & 0.016\%      \\
                      & $D_{US-20s-F}$ & $EER$           & 3.67\%        & 0.009\%      &           & $D_{US-30s-F}$ & $EER$           & 2.95\%        & 0.010\%      &           & $D_{US-60s-F}$ & $EER$           & 1.88\%        & 0.035\%      \\
                      &                &                 &               &              &           &                &                 &               &              &           &                &                 &               &              \\
                      & $D_{US-20s-M}$ & $FPR_{1}$       & 99.76\%       & 0.029\%      &           & $D_{US-30s-M}$ & $FPR_{1}$       & 99.74\%       & 0.008\%      &           & $D_{US-60s-M}$ & $FPR_{1}$       & 99.90\%       & 0.003\%      \\
                      & $D_{US-20s-F}$ & $FPR_{1}$       & 99.61\%       & 0.004\%      &           & $D_{US-30s-F}$ & $FPR_{1}$       & 99.69\%       & 0.004\%      &           & $D_{US-60s-F}$ & $FPR_{1}$       & 99.79\%       & 0.009\%      \\
D03     & $D_{US-20s-M}$ & $FPR_{2}$       & 99.66\%       & 0.021\%      &           & $D_{US-30s-M}$ & $FPR_{2}$       & 99.64\%       & 0.058\%      &           & $D_{US-60s-M}$ & $FPR_{2}$       & 99.87\%       & 0.003\%      \\
                      & $D_{US-20s-F}$ & $FPR_{2}$       & 99.52\%       & 0.007\%      &           & $D_{US-30s-F}$ & $FPR_{2}$       & 99.53\%       & 0.006\%      &           & $D_{US-60s-F}$ & $FPR_{2}$       & 99.74\%       & 0.018\%      \\
                      & $D_{US-20s-M}$ & $FPR_{3}$       & 99.86\%       & 0.015\%      &           & $D_{US-30s-M}$ & $FPR_{3}$       & 99.84\%       & 0.011\%      &           & $D_{US-60s-M}$ & $FPR_{3}$       & 99.95\%       & 0.000\%      \\
                      & $D_{US-20s-F}$ & $FPR_{3}$       & 99.79\%       & 0.004\%      &           & $D_{US-30s-F}$ & $FPR_{3}$       & 99.82\%       & 0.006\%      &           & $D_{US-60s-F}$ & $FPR_{3}$       & 99.87\%       & 0.017\%      \\
                      & $D_{US-20s-M}$ & $EER$           & 63.15\%       & 0.033\%      &           & $D_{US-30s-M}$ & $EER$           & 61.51\%       & 0.052\%      &           & $D_{US-60s-M}$ & $EER$           & 64.31\%       & 0.009\%      \\
                      & $D_{US-20s-F}$ & $EER$           & 60.19\%       & 0.009\%      &           & $D_{US-30s-F}$ & $EER$           & 60.49\%       & 0.005\%      &           & $D_{US-60s-F}$ & $EER$           & 62.29\%       & 0.018\%      \\
                      &                &                 &               &              &           &                &                 &               &              &           &                &                 &               &              \\
                      & $D_{US-20s-M}$ & $FPR_{1}$       & 74.93\%       & 0.132\%      &           & $D_{US-30s-M}$ & $FPR_{1}$       & 81.39\%       & 0.278\%      &           & $D_{US-60s-M}$ & $FPR_{1}$       & 76.29\%       & 0.051\%      \\
                      & $D_{US-20s-F}$ & $FPR_{1}$       & 52.45\%       & 0.024\%      &           & $D_{US-30s-F}$ & $FPR_{1}$       & 41.52\%       & 0.022\%      &           & $D_{US-60s-F}$ & $FPR_{1}$       & 64.68\%       & 0.125\%      \\
D04           & $D_{US-20s-M}$ & $FPR_{2}$       & 75.93\%       & 0.192\%      &           & $D_{US-30s-M}$ & $FPR_{2}$       & 81.99\%       & 0.248\%      &           & $D_{US-60s-M}$ & $FPR_{2}$       & 77.25\%       & 0.056\%      \\
                      & $D_{US-20s-F}$ & $FPR_{2}$       & 53.62\%       & 0.037\%      &           & $D_{US-30s-F}$ & $FPR_{2}$       & 42.47\%       & 0.051\%      &           & $D_{US-60s-F}$ & $FPR_{2}$       & 66.08\%       & 0.147\%      \\
                      & $D_{US-20s-M}$ & $FPR_{3}$       & 74.16\%       & 0.149\%      &           & $D_{US-30s-M}$ & $FPR_{3}$       & 80.83\%       & 0.240\%      &           & $D_{US-60s-M}$ & $FPR_{3}$       & 75.33\%       & 0.059\%      \\
                      & $D_{US-20s-F}$ & $FPR_{3}$       & 51.36\%       & 0.016\%      &           & $D_{US-30s-F}$ & $FPR_{3}$       & 40.49\%       & 0.050\%      &           & $D_{US-60s-F}$ & $FPR_{3}$       & 63.55\%       & 0.162\%      \\
                      & $D_{US-20s-M}$ & $EER$           & 27.96\%       & 0.086\%      &           & $D_{US-30s-M}$ & $EER$           & 33.98\%       & 0.236\%      &           & $D_{US-60s-M}$ & $EER$           & 26.83\%       & 0.027\%      \\
                      & $D_{US-20s-F}$ & $EER$           & 23.02\%       & 0.020\%      &           & $D_{US-30s-F}$ & $EER$           & 19.18\%       & 0.025\%      &           & $D_{US-60s-F}$ & $EER$           & 25.85\%       & 0.040\%      \\
                      &                &                 &               &              &           &                &                 &               &              &           &                &                 &               &              \\
                      & $D_{US-20s-M}$ & $FPR_{1}$       & 100.00\%      & 0.000\%      &           & $D_{US-30s-M}$ & $FPR_{1}$       & 100.00\%      & 0.000\%      &           & $D_{US-60s-M}$ & $FPR_{1}$       & 100.00\%      & 0.000\%      \\
                      & $D_{US-20s-F}$ & $FPR_{1}$       & 100.00\%      & 0.000\%      &           & $D_{US-30s-F}$ & $FPR_{1}$       & 100.00\%      & 0.000\%      &           & $D_{US-60s-F}$ & $FPR_{1}$       & 100.00\%      & 0.000\%      \\
D05       & $D_{US-20s-M}$ & $FPR_{2}$       & 100.00\%      & 0.000\%      &           & $D_{US-30s-M}$ & $FPR_{2}$       & 100.00\%      & 0.000\%      &           & $D_{US-60s-M}$ & $FPR_{2}$       & 100.00\%      & 0.000\%      \\
                      & $D_{US-20s-F}$ & $FPR_{2}$       & 100.00\%      & 0.000\%      &           & $D_{US-30s-F}$ & $FPR_{2}$       & 100.00\%      & 0.000\%      &           & $D_{US-60s-F}$ & $FPR_{2}$       & 100.00\%      & 0.000\%      \\
                      & $D_{US-20s-M}$ & $FPR_{3}$       & 100.00\%      & 0.000\%      &           & $D_{US-30s-M}$ & $FPR_{3}$       & 100.00\%      & 0.000\%      &           & $D_{US-60s-M}$ & $FPR_{3}$       & 100.00\%      & 0.000\%      \\
                      & $D_{US-20s-F}$ & $FPR_{3}$       & 100.00\%      & 0.000\%      &           & $D_{US-30s-F}$ & $FPR_{3}$       & 100.00\%      & 0.000\%      &           & $D_{US-60s-F}$ & $FPR_{3}$       & 100.00\%      & 0.000\%      \\
                      & $D_{US-20s-M}$ & $EER$           & 68.59\%       & 0.072\%      &           & $D_{US-30s-M}$ & $EER$           & 70.28\%       & 0.113\%      &           & $D_{US-60s-M}$ & $EER$           & 70.15\%       & 0.024\%      \\
                      & $D_{US-20s-F}$ & $EER$           & 66.56\%       & 0.005\%      &           & $D_{US-30s-F}$ & $EER$           & 69.57\%       & 0.022\%      &           & $D_{US-60s-F}$ & $EER$           & 72.71\%       & 0.034\%      \\
                      &                &                 &               &              &           &                &                 &               &              &           &                &                 &               &              \\
                      & $D_{US-20s-M}$ & $FPR_{1}$       & 85.37\%       & 0.097\%      &           & $D_{US-30s-M}$ & $FPR_{1}$       & 89.47\%       & 0.166\%      &           & $D_{US-60s-M}$ & $FPR_{1}$       & 72.08\%       & 0.117\%      \\
                      & $D_{US-20s-F}$ & $FPR_{1}$       & 85.78\%       & 0.020\%      &           & $D_{US-30s-F}$ & $FPR_{1}$       & 82.08\%       & 0.065\%      &           & $D_{US-60s-F}$ & $FPR_{1}$       & 80.66\%       & 0.135\%      \\
D06     & $D_{US-20s-M}$ & $FPR_{2}$       & 80.23\%       & 0.169\%      &           & $D_{US-30s-M}$ & $FPR_{2}$       & 85.61\%       & 0.237\%      &           & $D_{US-60s-M}$ & $FPR_{2}$       & 65.61\%       & 0.063\%      \\
                      & $D_{US-20s-F}$ & $FPR_{2}$       & 80.10\%       & 0.025\%      &           & $D_{US-30s-F}$ & $FPR_{2}$       & 74.87\%       & 0.021\%      &           & $D_{US-60s-F}$ & $FPR_{2}$       & 71.35\%       & 0.145\%      \\
                      & $D_{US-20s-M}$ & $FPR_{3}$       & 90.53\%       & 0.101\%      &           & $D_{US-30s-M}$ & $FPR_{3}$       & 93.79\%       & 0.166\%      &           & $D_{US-60s-M}$ & $FPR_{3}$       & 79.90\%       & 0.032\%      \\
                      & $D_{US-20s-F}$ & $FPR_{3}$       & 91.44\%       & 0.011\%      &           & $D_{US-30s-F}$ & $FPR_{3}$       & 88.88\%       & 0.039\%      &           & $D_{US-60s-F}$ & $FPR_{3}$       & 88.94\%       & 0.104\%      \\
                      & $D_{US-20s-M}$ & $EER$           & 43.51\%       & 0.112\%      &           & $D_{US-30s-M}$ & $EER$           & 46.43\%       & 0.135\%      &           & $D_{US-60s-M}$ & $EER$           & 36.06\%       & 0.062\%      \\
                      & $D_{US-20s-F}$ & $EER$           & 42.12\%       & 0.019\%      &           & $D_{US-30s-F}$ & $EER$           & 39.81\%       & 0.033\%      &           & $D_{US-60s-F}$ & $EER$           & 34.96\%       & 0.068\%     
                      \\
                      \bottomrule
                      \end{tabular}}
    \label{tab:sup_gender_bias}
\end{table*}

\begin{table*}
    \centering
    \caption{Absolute performance of detectors in male age bias study.}
    \resizebox{\textwidth}{!}{%
\begin{tabular}{@{}ll|ll|ll|ll|ll|ll|ll@{}}
\textbf{Method} & \textbf{} & \multicolumn{2}{c}{\textbf{teens}} & \multicolumn{2}{c}{\textbf{20s}} & \multicolumn{2}{c}{\textbf{30s}} & \multicolumn{2}{c}{\textbf{40s}} & \multicolumn{2}{c}{\textbf{50s}} & \multicolumn{2}{c}{\textbf{60s}} \\
\toprule
\textbf{}       & \textbf{} & \textbf{Mean}     & \textbf{SD}    & \textbf{Mean}    & \textbf{SD}   & \textbf{Mean}    & \textbf{SD}   & \textbf{Mean}    & \textbf{SD}   & \textbf{Mean}    & \textbf{SD}   & \textbf{Mean}    & \textbf{SD}   \\
\midrule
                & $FPR_{1}$ & 97.20\%           & 0.077\%        & 98.23\%          & 0.052\%       & 98.52\%          & 0.161\%       & 97.74\%          & 0.051\%       & 97.74\%          & 0.051\%       & 98.56\%          & 0.106\%       \\
TSSDNet         & $FPR_{2}$ & 99.95\%           & 0.022\%        & 99.96\%          & 0.026\%       & 99.99\%          & 0.008\%       & 99.94\%          & 0.011\%       & 99.93\%          & 0.013\%       & 99.92\%          & 0.015\%       \\
                & $FPR_{3}$ & 79.01\%           & 0.169\%        & 82.62\%          & 0.390\%       & 83.12\%          & 0.317\%       & 76.50\%          & 0.199\%       & 78.71\%          & 0.116\%       & 91.84\%          & 0.301\%       \\
                & $EER$     & 44.56\%           & 0.151\%        & 46.49\%          & 0.332\%       & 43.20\%          & 0.151\%       & 41.98\%          & 0.336\%       & 44.36\%          & 0.126\%       & 57.87\%          & 0.130\%       \\
                &           & \multicolumn{2}{c}{}               & \multicolumn{2}{c}{}             & \multicolumn{2}{c}{}             & \multicolumn{2}{c}{}             & \multicolumn{2}{c}{}             & \multicolumn{2}{c}{}             \\
                & $FPR_{1}$ & 33.59\%           & 0.353\%        & 27.01\%          & 0.271\%       & 21.71\%          & 0.176\%       & 26.03\%          & 0.307\%       & 27.01\%          & 0.215\%       & 28.61\%          & 0.341\%       \\
Wav2Vec2        & $FPR_{2}$ & 92.00\%           & 0.332\%        & 90.62\%          & 0.324\%       & 83.82\%          & 0.322\%       & 85.04\%          & 0.300\%       & 87.36\%          & 0.183\%       & 93.07\%          & 0.097\%       \\
                & $FPR_{3}$ & 4.09\%            & 0.154\%        & 2.63\%           & 0.128\%       & 2.35\%           & 0.184\%       & 4.03\%           & 0.182\%       & 3.23\%           & 0.134\%       & 1.47\%           & 0.073\%       \\
                & $EER$     & 5.07\%            & 0.045\%        & 3.79\%           & 0.086\%       & 3.68\%           & 0.120\%       & 4.90\%           & 0.164\%       & 4.37\%           & 0.051\%       & 2.97\%           & 0.096\%       \\
                &           & \multicolumn{2}{c}{}               & \multicolumn{2}{c}{}             & \multicolumn{2}{c}{}             & \multicolumn{2}{c}{}             & \multicolumn{2}{c}{}             & \multicolumn{2}{c}{}             \\
                & $FPR_{1}$ & 99.91\%           & 0.017\%        & 99.76\%          & 0.064\%       & 99.75\%          & 0.045\%       & 99.46\%          & 0.060\%       & 99.82\%          & 0.024\%       & 99.89\%          & 0.028\%       \\
Spec-ResNet     & $FPR_{2}$ & 99.86\%           & 0.029\%        & 99.67\%          & 0.088\%       & 99.65\%          & 0.061\%       & 99.31\%          & 0.061\%       & 99.73\%          & 0.032\%       & 99.88\%          & 0.024\%       \\
                & $FPR_{3}$ & 99.95\%           & 0.024\%        & 99.85\%          & 0.029\%       & 99.84\%          & 0.038\%       & 99.74\%          & 0.042\%       & 99.89\%          & 0.006\%       & 99.96\%          & 0.017\%       \\
                & $EER$     & 63.39\%           & 0.105\%        & 63.15\%          & 0.072\%       & 61.46\%          & 0.130\%       & 61.13\%          & 0.122\%       & 62.10\%          & 0.051\%       & 64.32\%          & 0.051\%       \\
                &           & \multicolumn{2}{c}{}               & \multicolumn{2}{c}{}             & \multicolumn{2}{c}{}             & \multicolumn{2}{c}{}             & \multicolumn{2}{c}{}             & \multicolumn{2}{c}{}             \\
                & $FPR_{1}$ & 69.56\%           & 0.547\%        & 75.08\%          & 0.551\%       & 81.67\%          & 0.298\%       & 79.80\%          & 0.493\%       & 80.77\%          & 0.237\%       & 76.26\%          & 0.180\%       \\
PS3DT           & $FPR_{2}$ & 70.33\%           & 0.423\%        & 75.91\%          & 0.408\%       & 82.20\%          & 0.347\%       & 80.82\%          & 0.364\%       & 81.76\%          & 0.083\%       & 77.48\%          & 0.454\%       \\
                & $FPR_{3}$ & 68.52\%           & 0.292\%        & 73.80\%          & 0.789\%       & 80.86\%          & 0.297\%       & 79.51\%          & 0.205\%       & 79.69\%          & 0.171\%       & 75.43\%          & 0.202\%       \\
                & $EER$     & 27.01\%           & 0.173\%        & 28.04\%          & 0.110\%       & 33.91\%          & 0.186\%       & 28.20\%          & 0.242\%       & 29.81\%          & 0.082\%       & 26.77\%          & 0.073\%       \\
                &           & \multicolumn{2}{c}{}               & \multicolumn{2}{c}{}             & \multicolumn{2}{c}{}             & \multicolumn{2}{c}{}             & \multicolumn{2}{c}{}             & \multicolumn{2}{c}{}             \\
                & $FPR_{1}$ & 100.00\%          & 0.000\%        & 100.00\%         & 0.000\%       & 100.00\%         & 0.000\%       & 100.00\%         & 0.000\%       & 100.00\%         & 0.000\%       & 100.00\%         & 0.000\%       \\
LFCC-GMMs        & $FPR_{2}$ & 100.00\%          & 0.000\%        & 100.00\%         & 0.000\%       & 100.00\%         & 0.000\%       & 100.00\%         & 0.000\%       & 100.00\%         & 0.000\%       & 100.00\%         & 0.000\%       \\
                & $FPR_{3}$ & 100.00\%          & 0.000\%        & 100.00\%         & 0.000\%       & 100.00\%         & 0.000\%       & 100.00\%         & 0.000\%       & 100.00\%         & 0.000\%       & 100.00\%         & 0.000\%       \\
                & $EER$     & 67.14\%           & 0.117\%        & 68.57\%          & 0.178\%       & 70.33\%          & 0.059\%       & 68.21\%          & 0.122\%       & 69.04\%          & 0.074\%       & 70.21\%          & 0.117\%       \\
                &           &                   &                &                  &               &                  &               &                  &               &                  &               &                  &               \\
                & $FPR_{1}$ & 87.08\%           & 0.270\%        & 85.10\%          & 0.299\%       & 89.78\%          & 0.399\%       & 84.20\%          & 0.256\%       & 83.22\%          & 0.120\%       & 71.81\%          & 0.316\%       \\
MFCC-ResNet     & $FPR_{2}$ & 81.88\%           & 0.331\%        & 80.14\%          & 0.355\%       & 85.36\%          & 0.255\%       & 78.78\%          & 0.608\%       & 76.96\%          & 0.289\%       & 65.66\%          & 0.312\%       \\
                & $FPR_{3}$ & 92.06\%           & 0.231\%        & 90.48\%          & 0.225\%       & 93.55\%          & 0.169\%       & 89.73\%          & 0.356\%       & 88.89\%          & 0.180\%       & 79.84\%          & 0.232\%       \\
                & $EER$     & 43.94\%           & 0.122\%        & 43.51\%          & 0.263\%       & 46.27\%          & 0.084\%       & 43.27\%          & 0.168\%       & 40.52\%          & 0.174\%       & 35.93\%          & 0.139\%      
\\
\bottomrule
\end{tabular}}
    \label{tab:sup_age_m}
\end{table*}

\begin{table*}
    \centering
    \caption{Absolute performance of detectors in female age bias study.}
    \resizebox{\textwidth}{!}{%
\begin{tabular}{@{}ll|ll|ll|ll|ll|ll|ll@{}}
\textbf{Method} & \textbf{} & \multicolumn{2}{c}{\textbf{teens}} & \multicolumn{2}{c}{\textbf{20s}} & \multicolumn{2}{c}{\textbf{30s}} & \multicolumn{2}{c}{\textbf{40s}} & \multicolumn{2}{c}{\textbf{50s}} & \multicolumn{2}{c}{\textbf{60s}} \\
\toprule
\textbf{}       & \textbf{} & \textbf{Mean}     & \textbf{SD}    & \textbf{Mean}    & \textbf{SD}   & \textbf{Mean}    & \textbf{SD}   & \textbf{Mean}    & \textbf{SD}   & \textbf{Mean}    & \textbf{SD}   & \textbf{Mean}    & \textbf{SD}   \\
\midrule
                & $FPR_{1}$ & 98.45\%                            & 0.088\%                       & 96.79\%                          & 0.241\%                       & 97.69\%                          & 0.055\%                       & 97.66\%                          & 0.025\%                       & 95.88\%                          & 0.063\%                       & 91.91\%                          & 0.345\%                       \\
TSSDNet         & $FPR_{2}$ & 99.91\%                            & 0.015\%                       & 99.61\%                          & 0.074\%                       & 99.88\%                          & 0.019\%                       & 99.80\%                          & 0.005\%                       & 99.84\%                          & 0.019\%                       & 99.93\%                          & 0.018\%                       \\
                & $FPR_{3}$ & 83.16\%                            & 0.230\%                       & 81.91\%                          & 0.334\%                       & 77.49\%                          & 0.245\%                       & 63.33\%                          & 0.061\%                       & 73.85\%                          & 0.298\%                       & 65.90\%                          & 0.245\%                       \\
                & $EER$     & 44.22\%                            & 0.227\%                       & 45.12\%                          & 0.213\%                       & 44.39\%                          & 0.119\%                       & 38.30\%                          & 0.035\%                       & 42.70\%                          & 0.121\%                       & 43.03\%                          & 0.069\%                       \\
                &           & \multicolumn{1}{c}{}               & \multicolumn{1}{c}{}          & \multicolumn{1}{c}{}             & \multicolumn{1}{c}{}          & \multicolumn{1}{c}{}             & \multicolumn{1}{c}{}          & \multicolumn{1}{c}{}             & \multicolumn{1}{c}{}          & \multicolumn{1}{c}{}             & \multicolumn{1}{c}{}          & \multicolumn{1}{c}{}             & \multicolumn{1}{c}{}          \\
                & $FPR_{1}$ & 35.85\%                            & 0.325\%                       & 29.53\%                          & 0.430\%                       & 23.79\%                          & 0.284\%                       & 30.43\%                          & 0.028\%                       & 23.24\%                          & 0.216\%                       & 17.37\%                          & 0.324\%                       \\
Wav2Vec2        & $FPR_{2}$ & 92.46\%                            & 0.178\%                       & 91.52\%                          & 0.128\%                       & 90.04\%                          & 0.265\%                       & 94.16\%                          & 0.012\%                       & 88.75\%                          & 0.224\%                       & 92.79\%                          & 0.207\%                       \\
                & $FPR_{3}$ & 12.78\%                            & 0.229\%                       & 2.34\%                           & 0.180\%                       & 1.38\%                           & 0.067\%                       & 1.87\%                           & 0.015\%                       & 2.70\%                           & 0.117\%                       & 0.63\%                           & 0.041\%                       \\
                & $EER$     & 11.53\%                            & 0.152\%                       & 3.62\%                           & 0.104\%                       & 2.88\%                           & 0.117\%                       & 3.21\%                           & 0.011\%                       & 3.97\%                           & 0.059\%                       & 1.92\%                           & 0.073\%                       \\
                &           & \multicolumn{1}{c}{}               & \multicolumn{1}{c}{}          & \multicolumn{1}{c}{}             & \multicolumn{1}{c}{}          & \multicolumn{1}{c}{}             & \multicolumn{1}{c}{}          & \multicolumn{1}{c}{}             & \multicolumn{1}{c}{}          & \multicolumn{1}{c}{}             & \multicolumn{1}{c}{}          & \multicolumn{1}{c}{}             & \multicolumn{1}{c}{}          \\
                & $FPR_{1}$ & 99.80\%                            & 0.037\%                       & 99.64\%                          & 0.090\%                       & 99.68\%                          & 0.040\%                       & 99.70\%                          & 0.005\%                       & 99.00\%                          & 0.063\%                       & 99.78\%                          & 0.045\%                       \\
Spec-ResNet     & $FPR_{2}$ & 99.70\%                            & 0.046\%                       & 99.53\%                          & 0.034\%                       & 99.50\%                          & 0.024\%                       & 99.62\%                          & 0.000\%                       & 98.80\%                          & 0.031\%                       & 99.76\%                          & 0.053\%                       \\
                & $FPR_{3}$ & 99.89\%                            & 0.013\%                       & 99.80\%                          & 0.029\%                       & 99.81\%                          & 0.027\%                       & 99.84\%                          & 0.000\%                       & 99.54\%                          & 0.071\%                       & 99.87\%                          & 0.027\%                       \\
                & $EER$     & 58.91\%                            & 0.069\%                       & 60.14\%                          & 0.063\%                       & 60.50\%                          & 0.099\%                       & 61.52\%                          & 0.017\%                       & 59.13\%                          & 0.142\%                       & 62.33\%                          & 0.075\%                       \\
                &           & \multicolumn{1}{c}{}               & \multicolumn{1}{c}{}          & \multicolumn{1}{c}{}             & \multicolumn{1}{c}{}          & \multicolumn{1}{c}{}             & \multicolumn{1}{c}{}          & \multicolumn{1}{c}{}             & \multicolumn{1}{c}{}          & \multicolumn{1}{c}{}             & \multicolumn{1}{c}{}          & \multicolumn{1}{c}{}             & \multicolumn{1}{c}{}          \\
                & $FPR_{1}$ & 62.05\%                            & 0.212\%                       & 52.87\%                          & 0.299\%                       & 41.59\%                          & 0.312\%                       & 55.38\%                          & 0.015\%                       & 42.80\%                          & 0.175\%                       & 64.47\%                          & 0.505\%                       \\
PS3DT           & $FPR_{2}$ & 62.58\%                            & 0.302\%                       & 53.60\%                          & 0.245\%                       & 42.29\%                          & 0.256\%                       & 56.77\%                          & 0.057\%                       & 44.17\%                          & 0.419\%                       & 65.95\%                          & 0.173\%                       \\
                & $FPR_{3}$ & 61.27\%                            & 0.394\%                       & 51.55\%                          & 0.551\%                       & 40.47\%                          & 0.330\%                       & 53.91\%                          & 0.055\%                       & 41.68\%                          & 0.588\%                       & 63.62\%                          & 0.285\%                       \\
                & $EER$     & 30.58\%                            & 0.177\%                       & 22.97\%                          & 0.127\%                       & 19.18\%                          & 0.170\%                       & 20.66\%                          & 0.015\%                       & 18.08\%                          & 0.027\%                       & 25.74\%                          & 0.140\%                       \\
                &           & \multicolumn{1}{c}{}               & \multicolumn{1}{c}{}          & \multicolumn{1}{c}{}             & \multicolumn{1}{c}{}          & \multicolumn{1}{c}{}             & \multicolumn{1}{c}{}          & \multicolumn{1}{c}{}             & \multicolumn{1}{c}{}          & \multicolumn{1}{c}{}             & \multicolumn{1}{c}{}          & \multicolumn{1}{c}{}             & \multicolumn{1}{c}{}          \\
                & $FPR_{1}$ & 100.00\%                           & 0.000\%                       & 100.00\%                         & 0.000\%                       & 100.00\%                         & 0.000\%                       & 100.00\%                         & 0.000\%                       & 100.00\%                         & 0.000\%                       & 100.00\%                         & 0.000\%                       \\
LFCC-GMMs        & $FPR_{2}$ & 100.00\%                           & 0.000\%                       & 100.00\%                         & 0.000\%                       & 100.00\%                         & 0.000\%                       & 100.00\%                         & 0.000\%                       & 100.00\%                         & 0.000\%                       & 100.00\%                         & 0.000\%                       \\
                & $FPR_{3}$ & 100.00\%                           & 0.000\%                       & 100.00\%                         & 0.000\%                       & 100.00\%                         & 0.000\%                       & 100.00\%                         & 0.000\%                       & 100.00\%                         & 0.000\%                       & 100.00\%                         & 0.000\%                       \\
                & $EER$     & 65.89\%                            & 0.058\%                       & 66.47\%                          & 0.072\%                       & 69.57\%                          & 0.141\%                       & 67.87\%                          & 0.012\%                       & 68.27\%                          & 0.104\%                       & 72.64\%                          & 0.117\%                       \\
                &           &                                    &                               &                                  &                               &                                  &                               &                                  &                               &                                  &                               &                                  &                               \\
                & $FPR_{1}$ & 89.20\%                            & 0.091\%                       & 85.84\%                          & 0.210\%                       & 81.96\%                          & 0.284\%                       & 80.44\%                          & 0.012\%                       & 77.84\%                          & 0.355\%                       & 80.58\%                          & 0.359\%                       \\
MFCC-ResNet     & $FPR_{2}$ & 84.37\%                            & 0.266\%                       & 80.16\%                          & 0.311\%                       & 75.04\%                          & 0.270\%                       & 73.22\%                          & 0.051\%                       & 70.16\%                          & 0.306\%                       & 70.98\%                          & 0.351\%                       \\
                & $FPR_{3}$ & 93.37\%                            & 0.081\%                       & 91.60\%                          & 0.248\%                       & 88.94\%                          & 0.195\%                       & 87.87\%                          & 0.036\%                       & 85.39\%                          & 0.088\%                       & 89.02\%                          & 0.204\%                       \\
                & $EER$     & 44.66\%                            & 0.102\%                       & 42.08\%                          & 0.194\%                       & 39.74\%                          & 0.110\%                       & 38.88\%                          & 0.017\%                       & 37.25\%                          & 0.167\%                       & 34.80\%                          & 0.272\%                      
\\ \bottomrule
\end{tabular}
}
    \label{tab:sup_age_f}
\end{table*}

\begin{table*}
    \centering
    \caption{Absolute performance of detectors in male accent bias study.}
\resizebox{1.0\textwidth}{!}{
\begin{tabular}{@{}ll|ll|ll|ll|ll|ll@{}}
\textbf{Method}      &           & \multicolumn{2}{c}{\textbf{Canadian}} & \multicolumn{2}{c}{\textbf{US}} & \multicolumn{2}{c}{\textbf{British}} & \multicolumn{2}{c}{\textbf{Australian}} & \multicolumn{2}{c}{\textbf{South Asian}} \\
\toprule
            &           & \textbf{Mean}          & \textbf{SD}           & \textbf{Mean}       & \textbf{SD}        & \textbf{Mean}          & \textbf{SD}          & \textbf{Mean}           & \textbf{SD}            & \textbf{Mean}            & \textbf{SD}            \\
\midrule
            & $FPR_{1}$ & 98.61\%       & 0.068\%      & 98.29\%    & 0.179\%   & 98.19\%       & 0.177\%     & 97.90\%        & 0.014\%       & 98.63\%         & 0.105\%       \\
TSSDNet     & $FPR_{2}$ & 99.99\%       & 0.006\%      & 99.94\%    & 0.010\%   & 99.96\%       & 0.018\%     & 100.00\%       & 0.000\%       & 99.95\%         & 0.010\%       \\
            & $FPR_{3}$ & 84.20\%       & 0.141\%      & 82.80\%    & 0.604\%   & 78.04\%       & 0.352\%     & 80.11\%        & 0.036\%       & 87.68\%         & 0.288\%       \\
            & $EER$     & 48.31\%       & 0.120\%      & 46.72\%    & 0.274\%   & 43.10\%       & 0.193\%     & 48.51\%        & 0.025\%       & 52.27\%         & 0.283\%       \\
            &           &               &              &            &           &               &             &                &               &                 &               \\
            & $FPR_{1}$ & 27.40\%       & 0.211\%      & 27.08\%    & 0.450\%   & 26.84\%       & 0.230\%     & 21.75\%        & 0.032\%       & 55.06\%         & 0.455\%       \\
Wav2Vec2    & $FPR_{2}$ & 91.04\%       & 0.207\%      & 90.68\%    & 0.333\%   & 90.91\%       & 0.148\%     & 88.13\%        & 0.027\%       & 95.45\%         & 0.117\%       \\
            & $FPR_{3}$ & 1.92\%        & 0.046\%      & 2.71\%     & 0.236\%   & 2.76\%        & 0.224\%     & 0.97\%         & 0.007\%       & 6.83\%          & 0.108\%       \\
            & $EER$     & 3.58\%        & 0.023\%      & 3.75\%     & 0.156\%   & 4.02\%        & 0.173\%     & 2.52\%         & 0.013\%       & 7.32\%          & 0.111\%       \\
            &           &               &              &            &           &               &             &                &               &                 &               \\
            & $FPR_{1}$ & 99.87\%       & 0.006\%      & 99.73\%    & 0.075\%   & 99.84\%       & 0.046\%     & 99.88\%        & 0.000\%       & 99.75\%         & 0.058\%       \\
Spec-Resnet & $FPR_{2}$ & 99.77\%       & 0.015\%      & 99.64\%    & 0.061\%   & 99.79\%       & 0.053\%     & 99.81\%        & 0.000\%       & 99.63\%         & 0.075\%       \\
            & $FPR_{3}$ & 99.92\%       & 0.007\%      & 99.86\%    & 0.026\%   & 99.91\%       & 0.032\%     & 99.90\%        & 0.000\%       & 99.86\%         & 0.049\%       \\
            & $EER$     & 63.92\%       & 0.035\%      & 63.09\%    & 0.064\%   & 63.39\%       & 0.063\%     & 63.79\%        & 0.007\%       & 62.13\%         & 0.090\%       \\
            &           &               &              &            &           &               &             &                &               &                 &               \\
            & $FPR_{1}$ & 71.78\%       & 0.076\%      & 75.12\%    & 0.789\%   & 83.29\%       & 0.200\%     & 81.76\%        & 0.042\%       & 77.87\%         & 0.279\%       \\
PS3DT       & $FPR_{2}$ & 72.92\%       & 0.181\%      & 75.20\%    & 0.258\%   & 84.05\%       & 0.175\%     & 82.62\%        & 0.040\%       & 78.41\%         & 0.400\%       \\
            & $FPR_{3}$ & 70.78\%       & 0.251\%      & 74.10\%    & 0.575\%   & 82.52\%       & 0.226\%     & 81.07\%        & 0.054\%       & 76.46\%         & 0.182\%       \\
            & $EER$     & 27.57\%       & 0.182\%      & 27.96\%    & 0.282\%   & 28.97\%       & 0.238\%     & 30.55\%        & 0.016\%       & 30.06\%         & 0.169\%       \\
            &           &               &              &            &           &               &             &                &               &                 &               \\
            & $FPR_{1}$ & 100.00\%      & 0.000\%      & 100.00\%   & 0.000\%   & 100.00\%      & 0.000\%     & 100.00\%       & 0.000\%       & 100.00\%        & 0.000\%       \\
LFCC-GMMs    & $FPR_{2}$ & 100.00\%      & 0.000\%      & 100.00\%   & 0.000\%   & 100.00\%      & 0.000\%     & 100.00\%       & 0.000\%       & 100.00\%        & 0.000\%       \\
            & $FPR_{3}$ & 100.00\%      & 0.000\%      & 100.00\%   & 0.000\%   & 100.00\%      & 0.000\%     & 100.00\%       & 0.000\%       & 100.00\%        & 0.007\%       \\
            & $EER$     & 68.52\%       & 0.058\%      & 68.68\%    & 0.224\%   & 67.74\%       & 0.130\%     & 69.95\%        & 0.008\%       & 63.47\%         & 0.124\%       \\
            &           & \multicolumn{2}{c}{}         & \multicolumn{2}{c}{}   & \multicolumn{2}{c}{}        & \multicolumn{2}{c}{}           & \multicolumn{2}{c}{}            \\
            & $FPR_{1}$ & 87.37\%       & 0.185\%      & 85.46\%    & 0.331\%   & 90.31\%       & 0.286\%     & 85.10\%        & 0.039\%       & 90.40\%         & 0.212\%       \\
MFCC-Resnet & $FPR_{2}$ & 82.22\%       & 0.270\%      & 80.38\%    & 0.485\%   & 85.85\%       & 0.294\%     & 79.47\%        & 0.051\%       & 85.71\%         & 0.287\%       \\
            & $FPR_{3}$ & 92.34\%       & 0.137\%      & 90.24\%    & 0.517\%   & 94.50\%       & 0.213\%     & 90.72\%        & 0.022\%       & 94.42\%         & 0.197\%       \\
            & $EER$     & 44.94\%       & 0.116\%      & 43.57\%    & 0.117\%   & 46.02\%       & 0.105\%     & 42.87\%        & 0.023\%       & 45.96\%         & 0.153\%      
\\ \bottomrule
\end{tabular}
}
    \label{tab:sup_accent_m}
\end{table*}
\begin{table*}
    \centering
    \caption{Absolute performance of detectors in female accent bias study.}
    \resizebox{\textwidth}{!}{
\begin{tabular}{@{}ll|ll|ll|ll|ll|ll@{}}
\textbf{Method}      &           & \multicolumn{2}{c}{\textbf{Canadian}} & \multicolumn{2}{c}{\textbf{US}} & \multicolumn{2}{c}{\textbf{British}} & \multicolumn{2}{c}{\textbf{Australian}} & \multicolumn{2}{c}{\textbf{South Asian}} \\
\toprule
            &           & \textbf{Mean}          & \textbf{SD}           & \textbf{Mean}       & \textbf{SD}        & \textbf{Mean}          & \textbf{SD}          & \textbf{Mean}           & \textbf{SD}            & \textbf{Mean}            & \textbf{SD}            \\
\midrule
            & $FPR_{1}$ & 95.05\%       & 0.034\%      & 96.73\%    & 0.156\%   & 97.84\%       & 0.062\%     & 99.22\%        & 0.049\%       & 99.02\%         & 0.088\%       \\
TSSDNet     & $FPR_{2}$ & 99.25\%       & 0.009\%      & 99.58\%    & 0.037\%   & 99.90\%       & 0.000\%     & 99.98\%        & 0.009\%       & 99.89\%         & 0.055\%       \\
            & $FPR_{3}$ & 80.24\%       & 0.077\%      & 81.74\%    & 0.308\%   & 77.77\%       & 0.215\%     & 91.80\%        & 0.107\%       & 91.36\%         & 0.286\%       \\
            & $EER$     & 45.07\%       & 0.062\%      & 45.19\%    & 0.289\%   & 43.12\%       & 0.027\%     & 60.72\%        & 0.135\%       & 49.14\%         & 0.181\%       \\
            &           &               &              &            &           &               &             &                &               &                 &               \\
            & $FPR_{1}$ & 21.50\%       & 0.049\%      & 29.28\%    & 0.419\%   & 33.89\%       & 0.057\%     & 61.50\%        & 0.201\%       & 73.76\%         & 0.593\%       \\
Wav2Vec2    & $FPR_{2}$ & 89.62\%       & 0.074\%      & 91.48\%    & 0.169\%   & 91.62\%       & 0.073\%     & 97.13\%        & 0.069\%       & 98.62\%         & 0.130\%       \\
            & $FPR_{3}$ & 0.91\%        & 0.011\%      & 2.31\%     & 0.101\%   & 4.67\%        & 0.065\%     & 2.97\%         & 0.096\%       & 27.53\%         & 0.618\%       \\
            & $EER$     & 2.31\%        & 0.000\%      & 3.63\%     & 0.203\%   & 5.52\%        & 0.054\%     & 4.70\%         & 0.061\%       & 18.61\%         & 0.389\%       \\
            &           &               &              &            &           &               &             &                &               &                 &               \\
            & $FPR_{1}$ & 99.66\%       & 0.009\%      & 99.56\%    & 0.049\%   & 99.74\%       & 0.011\%     & 99.84\%        & 0.020\%       & 98.78\%         & 0.052\%       \\
Spec-Resnet & $FPR_{2}$ & 99.52\%       & 0.018\%      & 99.49\%    & 0.102\%   & 99.61\%       & 0.009\%     & 99.84\%        & 0.017\%       & 98.58\%         & 0.102\%       \\
            & $FPR_{3}$ & 99.84\%       & 0.009\%      & 99.80\%    & 0.050\%   & 99.82\%       & 0.009\%     & 99.94\%        & 0.009\%       & 99.34\%         & 0.116\%       \\
            & $EER$     & 59.69\%       & 0.021\%      & 60.21\%    & 0.128\%   & 59.53\%       & 0.052\%     & 63.66\%        & 0.040\%       & 57.34\%         & 0.298\%       \\
            &           &               &              &            &           &               &             &                &               &                 &               \\
            & $FPR_{1}$ & 42.60\%       & 0.082\%      & 52.48\%    & 0.466\%   & 62.85\%       & 0.211\%     & 79.54\%        & 0.117\%       & 67.80\%         & 0.554\%       \\
PS3DT       & $FPR_{2}$ & 43.29\%       & 0.057\%      & 53.50\%    & 0.969\%   & 63.81\%       & 0.180\%     & 80.36\%        & 0.137\%       & 69.09\%         & 0.465\%       \\
            & $FPR_{3}$ & 41.85\%       & 0.062\%      & 51.43\%    & 0.463\%   & 61.47\%       & 0.057\%     & 78.63\%        & 0.266\%       & 66.57\%         & 0.454\%       \\
            & $EER$     & 20.02\%       & 0.025\%      & 22.73\%    & 0.211\%   & 27.63\%       & 0.075\%     & 26.95\%        & 0.040\%       & 28.36\%         & 0.333\%       \\
            &           &               &              &            &           &               &             &                &               &                 &               \\
            & $FPR_{1}$ & 100.00\%      & 0.000\%      & 100.00\%   & 0.000\%   & 100.00\%      & 0.000\%     & 100.00\%       & 0.000\%       & 100.00\%        & 0.000\%       \\
LFCC-GMMs    & $FPR_{2}$ & 100.00\%      & 0.000\%      & 100.00\%   & 0.000\%   & 100.00\%      & 0.000\%     & 100.00\%       & 0.000\%       & 100.00\%        & 0.000\%       \\
            & $FPR_{3}$ & 100.00\%      & 0.000\%      & 100.00\%   & 0.000\%   & 99.98\%       & 0.000\%     & 100.00\%       & 0.000\%       & 100.00\%        & 0.000\%       \\
            & $EER$     & 67.88\%       & 0.020\%      & 66.67\%    & 0.166\%   & 68.36\%       & 0.082\%     & 67.90\%        & 0.058\%       & 56.90\%         & 0.085\%       \\
            &           & \multicolumn{2}{c}{}         & \multicolumn{2}{c}{}   & \multicolumn{2}{c}{}        & \multicolumn{2}{c}{}           & \multicolumn{2}{c}{}            \\
            & $FPR_{1}$ & 86.47\%       & 0.014\%      & 85.82\%    & 0.475\%   & 80.61\%       & 0.148\%     & 91.32\%        & 0.072\%       & 89.29\%         & 0.188\%       \\
MFCC-Resnet & $FPR_{2}$ & 80.60\%       & 0.077\%      & 80.01\%    & 0.538\%   & 74.37\%       & 0.063\%     & 85.58\%        & 0.173\%       & 83.87\%         & 0.639\%       \\
            & $FPR_{3}$ & 92.74\%       & 0.049\%      & 91.27\%    & 0.223\%   & 86.93\%       & 0.106\%     & 95.44\%        & 0.087\%       & 93.93\%         & 0.309\%       \\
            & $EER$     & 42.34\%       & 0.039\%      & 42.37\%    & 0.189\%   & 40.75\%       & 0.063\%     & 45.45\%        & 0.138\%       & 43.44\%         & 0.269\%      
\\
\bottomrule
\end{tabular}
}
    \label{tab:sup_accent_f}
\end{table*}
\section{Obtaining $\Delta \textbf{FPR}$ and {$\Delta \textbf{EER}$}}\label{sec:sup_delta}
In this section, we describe how we obtain $\Delta \textbf{FPR}$ and $\Delta \textbf{EER}$ reported in the paper from the absolute value of $FPR$ and $EER$ reported in ~\cref{sec:sup_result}.
We will use example of age bias study for male speakers \ie ~\cref{tab:sup_age_m} to inform about our calculations.
Notice there are 6 different age groups. 
For each metric, we first obtained the minimum value. 
For example, $FPR_{1}$ for detector Wav2Vec2 has the minimum value for age group 30s in~\cref{tab:sup_age_m}. We refer to this value as $minFPR_{1}$ and then report $\Delta FPR_{1} := FPR_{1} - minFPR_{1}$.
Note the minimum will be different for each metric and detector.
We did this so that bias study and results are not dependent on individual detector performance and help to capture difference in performance by a detector on one age group versus another.
We use similar approach for calculating $\Delta FPR_{2}$, $\Delta FPR_{3}$, and $\Delta EER$.


\end{document}